\documentclass[lettersize,journal]{IEEEtran}
\usepackage{amsmath,amsfonts}
\usepackage{algorithmic}
\usepackage{algorithm}
\usepackage{array}
\usepackage[caption=false,font=normalsize,labelfont=sf,textfont=sf]{subfig}
\usepackage{textcomp}
\usepackage{stfloats}
\usepackage{url}
\usepackage{verbatim}
\usepackage{subcaption}   
\usepackage{graphicx}      
\usepackage{cite}
\usepackage{color}
\usepackage{bbding}
\usepackage{caption}
\usepackage{multirow}
\definecolor{R3}{rgb}{0.50,0.69,0.40}
\definecolor{yellow}{rgb}{0.835,0.711,0.336}

\usepackage[x11names,table]{xcolor}
\usepackage{booktabs}

\definecolor{lip}{HTML}{D79B00}
\definecolor{identity}{HTML}{9673A6}
\definecolor{expression}{HTML}{FF0000}
\definecolor{mouth}{HTML}{82B366}
\definecolor{pose}{HTML}{6C8EBF}

\definecolor{f1}{HTML}{000000}
\definecolor{f2}{HTML}{000000}
\definecolor{f3}{HTML}{000000}

\begin{document}

\title{EDTalk++: Full Disentanglement for \\Controllable Talking Head Synthesis}


\author{Shuai Tan, Bin Ji, Ye Pan\IEEEauthorrefmark{2}

\IEEEcompsocitemizethanks{
\IEEEcompsocthanksitem{\IEEEauthorrefmark{2}Corresponding author.}

\IEEEcompsocthanksitem Shuai Tan, Bin Ji and Ye Pan are with Department of Computer Science and Technology, Shanghai Jiao Tong University, China. E-mail: \{tanshuai0219, bin.ji, whitneypanye\}@sjtu.edu.cn.
}
\thanks{Preliminary versions of this work were published in ECCV 2024~\cite{tan2024edtalk}.}
}

\markboth{Journal of \LaTeX\ Class Files,~Vol.~14, No.~8, August~2021}%
{Shell \MakeLowercase{\textit{et al.}}: A Sample Article Using IEEEtran.cls for IEEE Journals}


\maketitle

\begin{abstract}
Achieving disentangled control over multiple facial motions and accommodating diverse input modalities greatly enhances the application and entertainment of the talking head generation. This necessitates a deep exploration of the decoupling space for facial features, ensuring that they \textbf{a)} operate independently without mutual interference and \textbf{b)} can be preserved to share with different modal inputs—both aspects often neglected in existing methods. To address this gap, this paper proposes \textbf{EDTalk++}, a novel full disentanglement framework for controllable talking head generation. Our framework enables individual manipulation of mouth shape, head pose, eye movement, and emotional expression, conditioned on video or audio inputs. Specifically, we employ four \textbf{lightweight} modules to decompose the facial dynamics into four distinct latent spaces representing mouth, pose, eye, and expression, respectively. Each space is characterized by a set of learnable bases whose linear combinations define specific motions. To ensure independence and accelerate training, we enforce orthogonality among bases and devise an \textbf{efficient} training strategy to allocate motion responsibilities to each space without relying on external knowledge. The learned bases are then stored in corresponding banks, enabling shared visual priors with audio input. Furthermore, considering the properties of each space, we propose an Audio-to-Motion module for audio-driven talking head synthesis. Experiments are conducted to demonstrate the effectiveness of EDTalk++.
\end{abstract}


\section{Introduction}
\IEEEPARstart{T}{alking} head animation~\cite{tan2024style2talker, ye1, ye2, ye3,guo2018cnn,wang2024styletalk++,hong2023dagan++,yu2024metaearth,tan2025fixtalk, ji2025pomp, tan2025animaPAMI,tan2025synmotion,wei2025dreamrelation,liu2025vqtalker} has garnered significant research attention owing to its wide-ranging applications in education, filmmaking, virtual digital humans, and the entertainment industry~\cite{pataranutaporn2021ai}. While previous methods~\cite{shen2022learning, khakhulin2022realistic, yang2022face2face, yin2022styleheat} have achieved notable advancements, most of them generate talking head videos in a holistic manner, lacking fine-grained individual control. Consequently, attaining precise and disentangled manipulation over various facial motions such as mouth shapes, head poses, eye movements, and emotional expressions remains a challenge, crucial for crafting lifelike avatars~\cite{wang2023progressive}. Moreover, existing approaches typically cater to only one driving source: either audio~\cite{liu2024anitalker,thies2020neural} or video~\cite{siarohin2019first, hong2022depth}, thereby limiting their applicability in the multimodal context. There is a pressing need for a unified framework capable of simultaneously achieving individual facial control and handling both audio-driven and video-driven talking face generation.

\begin{figure*}[t]
  \centering
\includegraphics[width=1\linewidth]{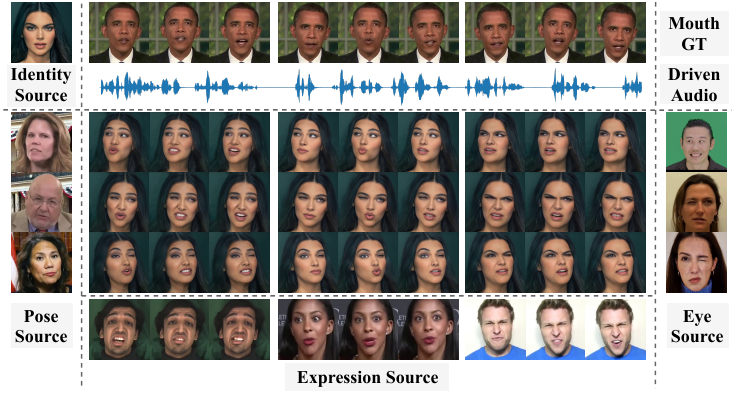}
    \caption{Illustrative animations produced by EDTalk++. Given an identity source, EDTalk++ synthesizes talking face videos characterized by mouth shapes, head poses, eye movements, and emotion expressions consistent with mouth GT, pose source, eye source and expression source. These facial dynamics can also be inferred directly from driven audio.}
    \label{fig:teaser}
\end{figure*}

To tackle the challenges, an intuition is to disentangle the entirety of facial dynamics into distinct facial latent spaces dedicated to individual components. However, it is non-trivial due to the intricate interplay among facial movements~\cite{wang2023progressive}. For instance, mouth shapes profoundly impact emotional expressions, where one speaks happily with upper lip corners but sadly with the depressed ones~\cite{gan2023efficient,ekman1978facial}. Despite the extensive efforts made in facial disentanglement by previous studies~\cite{yu2023talking, liang2022expressive, wang2023progressive, pang2023dpe, zhou2021pose}, we argue there exist three key limitations. \textbf{\textcolor{black}{(1)}} Overreliance on external and prior information increases the demand for data and complicates the data pre-processing: One popular line~\cite{yu2023talking, liang2022expressive, wang2023progressive} relies heavily on external audio data to decouple the mouth space via contrastive learning~\cite{khosla2020supervised}. Subsequently, they further disentangle the pose space using predefined 6D pose coefficients extracted from 3D face reconstruction models~\cite{danvevcek2022emoca}. However, such external and prior information escalates dataset demands and any inaccuracies therein can lead to the trained model errors. {\textbf{\textcolor{black}{(2)}} Disentangling latent spaces without internal constraints leads to incomplete decoupling.} Previous works~\cite{zhou2021pose, liang2022expressive} simply constrain each space externally with a prior during the decoupling process, overlooking inter-space constraints. This oversight fails to ensure that each space exclusively handles its designated component without interference from others, leading to training complexities, reduced efficiency, and performance degradation. {\textbf{\textcolor{black}{(3)}} Inefficient training strategy escalates the training time and computational cost.} When disentangling a new sub-space, some methods~\cite{pang2023dpe, wang2023progressive} require training the entire heavyweight network from scratch, which significantly incurs high time and computational costs~\cite{gan2023efficient}. It can be costly and unaffordable for many researchers. Furthermore, most methods are unable to utilize audio and video inputs simultaneously.

To cope with such issues, this paper proposes an \textbf{E}fficient \textbf{D}isentanglement framework, tailored for one-shot talking head generation with precise control over mouth shape, head pose, eye movement and emotional expression, conditioned on video or audio inputs. Our key insight lies in our requirements for decoupled space: \textbf{\textcolor{black}{(a)}} The decoupled spaces should be disjoint, which means each space captures solely the motion of its corresponding component without the interference from others. This also ensures that decoupling a new space will not affect the trained models, thereby avoiding the necessity of training from scratch. \textbf{\textcolor{black}{(b)}} Once the spaces are disentangled from video data to support video-driven paradigm, they should be stored to share with the audio inputs for further audio-driven setting. 

To this end, drawing inspiration from the observation that the entire motion space can be represented by a set of directions~\cite{wang2021latent}, we innovatively disentangle the whole motion space into four distinct component-aware latent spaces. Each space is characterized by a set of learnable bases. To ensure that different latent spaces do not interfere with each other, we constrain bases orthogonal to each other not only \textit{intra}-space~\cite{wang2021latent} but also \textit{inter}-space. To accomplish the disentanglement without prior information, we introduce a progressive training strategy comprising cross-reconstruction disentanglement which decouples the pose, mouth and eye from the talking head videos, and self-reconstruction complementary learning for expression decoupling. Despite comprising multiple stages, our decoupling process involves training only the proposed lightweight Latent Navigation modules, keeping the weights of other heavier modules fixed for efficient training.

To explicitly preserve the disentangled latent spaces, we store the base sets of disentangled spaces in the corresponding banks. These banks serve as repositories of prior bases essential for audio-driven talking head generation. Consequently, we introduce an Audio-to-Motion module designed to predict the weights of the mouth, pose, eyes and expression banks, respectively. Specifically, we employ an audio encoder to synchronize lip motions with the audio input. Given the non-deterministic nature of head motions and eye movements~\cite{zhang2023sadtalker}, we utilize diffusion model~\cite{DDPM, DDIM, animatex, mimir, ji2025sport} to generate probabilistic and realistic poses by sampling a Gaussian noise, guided by the rhythm of audio. Regarding expression, we aim to extract emotional cues from the audio~\cite{ji2021audio} and transcripts. It ensures that the generated talking head video aligns with the tone and context of audio, eliminating the need for additional expression references. In this way, our EDTalk++ enables talking face generation directly from the sole audio input. 


\textbf{Difference from our conference version.} This manuscript presents significant improvements over our previous conference version~\cite{tan2024edtalk}, introducing more fine-grained facial disentanglement and a deeper experimental analysis. \textbf{1)} We extend the original framework to enable more precise disentanglement and control of eye movements, allowing for more controllable talking head generation. \textbf{2)} We enhance the audio-to-motion pipeline by incorporating a diffusion model to better capture the dynamics of both pose and eye movement, leading to more diverse and realistic outputs while significantly improving inference speed, thus supporting real-time generation. \textbf{3)} We upgrade the video generation pipeline from $256 \times 256$ to $512 \times 512$ resolution, accompanied with a higher-quality training dataset to reduce rendering artifacts. \textbf{4)} We provide a more detailed experimental evaluation of the model architecture and its disentangled generation capabilities.

Our contributions are outlined as follows: \textbf{1)} We present EDTalk++, an efficient disentanglement framework enabling precise control over talking head synthesis concerning mouth shape, head pose, eye movement and emotional expression. \textbf{2)} By introducing orthogonal bases and an efficient training strategy, we successfully achieve complete decoupling of these four spaces. Leveraging the properties of each space, we implement Audio-to-Motion modules to facilitate audio-driven talking face generation. \textbf{3)} Extensive experiments demonstrate that our EDTalk++ surpasses the competing methods in both quantitative and qualitative evaluation.

\section{Related Work}

\subsection{Disentanglement on the face}
Facial dynamics typically involve coordinated movements such as head poses, mouth shapes, and emotional expressions in a global manner~\cite{tan2023emmn}, making their separate control challenging. Several works have been developed to address this issue. PC-AVS~\cite{zhou2021pose} employs contrastive learning to isolate the mouth space related to audio. Yet since similar pronunciations tend to correspond to the same mouth shape~\cite{li2023ae}, the constructed negative pairs in a mini-batch often include positive pairs and the number of negative pairs in the mini-batch is too small~\cite{he2020momentum}, both of which results in subpar results. Similarly, PD-FGC~\cite{wang2023progressive} and TH-PAD~\cite{yu2023talking} face analogous challenges in obtaining content-related mouth spaces. Although TH-PAD incorporates lip motion decorrelation loss to extract non-lip space, it still retains a coupled space where expressions and head poses are intertwined. This coupling results in randomly generated expressions co-occurring with head poses, compromising user-friendliness and content relevance. Despite the achievement of PD-FGC in decoupling facial details, its laborious coarse-to-fine disentanglement process consumes substantial computational resources and time. DPE~\cite{pang2023dpe} introduces a bidirectional cyclic training strategy to disentangle head pose and expression from talking head videos. However, it necessitates two generators to independently edit expression and pose sequentially, escalating computational resource consumption and runtime. In contrast, we propose an efficient decoupling approach to segregate faces into mouth, head pose, eye and expression components, readily controllable by different sources. Moreover, our method requires only a unified generator, and minimal additional resources are needed when exploring a new disentangled space.

\subsection{Audio-driven Talking Head Generation}
Audio-driven talking head generation~\cite{bregler2023video, liu2022semantic, emo, xu2024vasa, cui2024hallo2, cui2024hallo3, echomimicv2, lin2024cyberhost, loopy, mimaface, aniportrait} endeavors to animate images with accurate lip movements synchronized with input audio clips. Research in this area is predominantly categorized into two groups: intermediate representation based methods and reconstruction-based methods. Intermediate representation based methods~\cite{zhou2020makelttalk,chen2019hierarchical,das2020speech,zakharov2019few,zhong2023identity,wang2021audio2head,wang2022one,chen2020talking,yang2022face2face} typically consist of two sub-modules: one predicts intermediate representations from audio, and the other synthesizes photorealistic images from these representations. For instance, Das et al.\cite{das2020speech} employ landmarks as an intermediate representation, utilizing an audio-to-landmark module and a landmark-to-image module to connect audio inputs and video outputs. Yin et al.\cite{yin2022styleheat} extract 3DMM parameters~\cite{blanz1999morphable} to warp source images using predicted flow fields. However, obtaining such intermediate representations, like landmarks and 3D models, is laborious and time-consuming. Moreover, they often offer limited facial dynamics details, and training the two sub-modules separately can accumulate errors, leading to suboptimal performance. In contrast, our approach operates within a reconstruction-based framework~\cite{thies2020neural,shen2022learning,chen2018lip,Song_Zhu_Li_Wang_Qi_2019,zhou2019talking,chen2023vast,wang2023lipformer,shen2023difftalk}. It integrates features extracted by encoders from various modalities to reconstruct talking head videos in an end-to-end manner, alleviating the aforementioned issues. A notable example is Wav2Lip~\cite{prajwal2020lip}, which employs an audio encoder, an identity encoder, and an image decoder to generate precise lip movements. Similarly, Zhou et al.~\cite{zhou2021pose} incorporate an additional pose encoder for free pose control, yet disregard the nondeterministic nature of natural movement. To address this, we propose employing a probabilistic model to establish a distribution of non-verbal head motions. Additionally, none of the existing methods consider facial expressions, crucial for authentic talking head generation. Our approach aims to integrate facial expressions into the model to enhance the realism and authenticity of the generated talking heads. 

\subsection{Emotional Talking Head Generation}
Emotional talking head generation is gaining traction due to its wide-ranging applications and heightened entertainment potential. On the one hand, some studies~\cite{ji2021audio,SanjanaSinha2022EmotionControllableGT,wang2020mead,gan2023efficient} identify emotions using discrete emotion labels, albeit facing a challenge to generate controllable and fine-grained expressions. On the other hand, recent methodologies~\cite{liang2022expressive,ji2022eamm,ma2023styletalk,wang2023progressive} incorporate emotional images or videos as references to indicate desired expressions. Ji et al.\cite{ji2022eamm}, for instance, mask the mouth region of an emotional video and utilize the remaining upper face as an expression reference for emotional talking face generation. However, as mouth shape plays a crucial role in conveying emotion\cite{tan2023emmn}, they struggle to synthesize vivid expressions due to their failure to decouple expressions from the entire face. Thanks to our orthogonal base and efficient training strategy, we are capable of fully disentangling different motion spaces like mouth shape and emotional expression, thus achieving finely controlled talking head synthesis. Moreover, we also incorporate emotion contained within audio and transcripts. To the best of our knowledge, we are the first to achieve this goal, which automatically infers suitable expressions from audio tone and text, thereby generating consistent emotional talking face videos without relying on explicit image/video references.

\section{Methodology}

\begin{figure*}[t]
  \centering
  \includegraphics[width=1\linewidth]{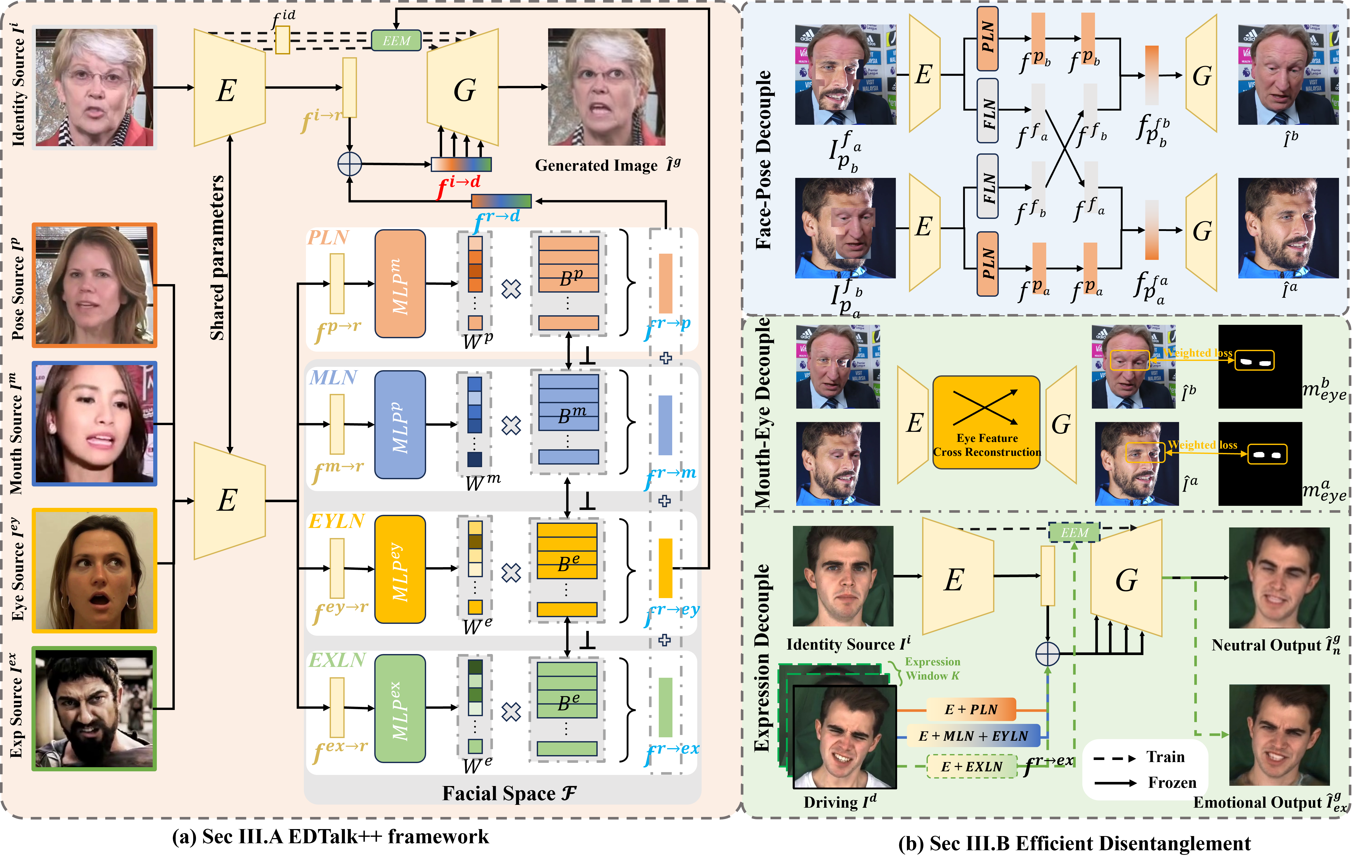}
    \caption{Illustration of our proposed EDTalk++. (a) EDTalk++ framework. Given an identity source $I^i$ and various driving images $I^*$ ($* \in \{m,p,ey, ex\}$) for controlling corresponding facial components, EDTalk++ animates the identity image $I^i$ to mimic the head pose, mouth shape, eye movement and expression of $I^p$, $I^m$, $I^{ey}$ and $I^{ex}$ with the assistance of four Component-aware Latent Navigation modules: PLN, MLN, EYLN and EXLN. (b) Efficient Disentanglement. The disentanglement process consists of three parts: Face-Pose Decouple, Mouth-Eye Decouple and Expression Decouple. For the first two stages, we introduce the cross-reconstruction training strategy aimed at separating head pose, mouth shape and eye movement. For the last stage, we achieve expression disentanglement using self-reconstruction complementary learning.}
    \label{fig:overview}
\end{figure*}

As illustrated in Fig.~\ref{fig:overview} (a), given an identity image $I^i$, we aim to synthesize emotional talking face image $\hat{I}^g$ that maintains consistency in identity information, mouth shape, head pose, eye movement, and emotional expression with various driving sources $I^i$, $I^m$, $I^p$, $I^{ey}$ and $I^{ex}$. Our intuition is to disentangle different facial components from the overall facial dynamics. To this end, we propose EDTalk++ (Sec. \ref{sec:framework}) with learnable orthogonal bases stored in banks $B^*$ ($*$ refers to the mouth source $m$, pose source $p$, eye source $ey$ and expression source $ex$ for simplicity), each representing a distinct direction of facial movements. To ensure the bases are component-aware, we propose an efficient disentanglement strategy (Sec. \ref{sec:disentanglement}), comprising Face-Pose Decoupling, Mouth-Eye Decoupling and Expression Decoupling, which decompose the overall facial motion into pose, mouth, eye, and expression spaces. Leveraging these disentangled spaces, we further explore an Audio-to-Motion module (Section~\ref{sec:audio_driven}, Figure~\ref{fig:audio_driven}) to produce audio-driven emotional talking face videos featuring probabilistic poses and eye movements, audio-synchronized lip motions, and semantically-aware expressions.

\subsection{EDTalk++ Framework}
\label{sec:framework}
Figure~\ref{fig:overview} (a) illustrates the structure of EDTalk++, which is based on an autoencoder architecture consisting of an Encoder $E$, four Component-aware Latent Navigation modules (CLNs) and a Generator $G$. The encoder $E$ maps the identity image $I^i$ and various driving source $I^*$ into the latent features $\textcolor{f1}{f^{i \rightarrow r}} = E(I^i)$ and $\textcolor{f1}{f^{* \rightarrow r}} = E(I^*)$. The process in inspired by FOMM~\cite{siarohin2019first} and LIA~\cite{wang2021latent}. Instead of directly modeling motion transformation \textcolor{f3}{$f^{i\rightarrow *}$} from identity image $I^i$ to driving image $I^*$ in the latent space, we posit the existence of a canonical feature $f^r$, that facilitates motion transfer between identity features and driving ones, expressed as $\textcolor{f3}{f^{i\rightarrow *}} = \textcolor{f1}{f^{i\rightarrow r}} + \textcolor{f2}{f^{r\rightarrow *}}$. 

Thus, upon acquiring the latent features \textcolor{f1}{$f^{* \rightarrow r}$} extracted by $E$ from driving images $I^*$, we devise four Component-aware Latent Navigation modules to transform them into $\textcolor{f2}{f^{r\rightarrow *}} = CLN(\textcolor{f1}{f^{* \rightarrow r}})$. For clarity, we use pose as an example, denoted as $*=p$. Within the Pose-aware Latent Navigation (PLN) module, we establish a pose bank $B^p = \{b^p_1, ..., b^p_n\}$ to store $n$ learnable base $b^p_i$. To ensure each base represents a distinct pose motion direction, we enforce orthogonality between every pair of bases by imposing a constraint of $\left\langle b^p_i, b^p_j \right \rangle= 0\quad (i \not= j)$, where $\left\langle \cdot, \cdot \right \rangle$ signifies the dot product operation. It allows us to depict various head pose movements as linear combinations of the bases. Consequently, we design a Multi-Layer Perceptron layer $MLP^p$ to predict the weights $W^p = \{w^p_1, ..., w^p_n\}$ of the pose bases from the latent feature $f^{p \rightarrow r}$:
\begin{equation}
    W^p = \{w^p_1, ..., w^p_n\} = MLP^p(\textcolor{f1}{f^{p \rightarrow r}})
\end{equation}
\begin{equation}
    \textcolor{f2}{f^{r \rightarrow p}} = \sum_{i=1}^{n} w^p_i b^p_i,
\end{equation}

Mouth, Eye and Expression-aware Latent Navigation module share the same architecture with PLN but have different parameters, where we can also derive $\textcolor{f2}{f^{r \rightarrow m}} = \sum_{i=1}^{n} w^m_i b^m_i, W^m = MLP^m(\textcolor{f1}{f^{m \rightarrow r}})$, $\textcolor{f2}{f^{r \rightarrow {ey}}} = \sum_{i=1}^{n} w^{ey}_i b^{ey}_i, W^{ey} = MLP^{ey}(\textcolor{f1}{f^{{ey} \rightarrow r}})$ and $\textcolor{f2}{f^{r \rightarrow {ex}}} = \sum_{i=1}^{n} w^{ex}_i b^{ex}_i, W^{ex} = MLP^{ex}(\textcolor{f1}{f^{{ex} \rightarrow r}})$ in the similar manner. It's worth noting that to achieve complete disentanglement of facial components and prevent changes in one component from affecting others, we ensure orthogonality between the four banks ($B^p,B^m,B^{ey},B^{ex}$). This also allows us to directly combine the four features to obtain the driving feature $\textcolor{f2}{f^{r \rightarrow d}} = \textcolor{f2}{f^{r \rightarrow p}}+\textcolor{f2}{f^{r \rightarrow m}}+\textcolor{f2}{f^{r \rightarrow ey}}+\textcolor{f2}{f^{r \rightarrow ex}}$. We further get $\textcolor{f3}{f^{i \rightarrow d}} = \textcolor{f1}{f^{i \rightarrow r}}+\textcolor{f2}{f^{r \rightarrow d}}$, which is subsequently fed into the Generator $G$ to synthesize the final result $\hat{I}^g$. To maintain identity information, $G$ incorporates the identity features $f^{id}$ of the identity image via skip connections. Additionally, to enhance emotional expressiveness with the assistance of the emotion feature $\textcolor{f2}{f^{r\rightarrow ex}}$, we introduce a lightweight plug-and-play Emotion Enhancement Module ($EEM$), which will be discussed in the subsequent subsection. In summary, the generation process can be formulated as follows:
\begin{equation}
\label{eq:g}
    \hat{I}^g = G(\textcolor{f3}{f^{i \rightarrow d}}, f^{id}, EEM(\textcolor{f2}{f^{r \rightarrow ex}})),
\end{equation}
where $EEM$ is exclusively utilized during emotional talking face generation. For brevity, we omit $f^{id}$ in the subsequent equations.

\subsection{Efficient Disentanglement}
\label{sec:disentanglement}
Based on the outlined framework, the crux lies in training each Component-aware Latent Navigation module to store only the bases corresponding to the motion of its respective components and to ensure no interference between different components. To achieve this, we approach facial disentanglement from the following two perspectives: \textbf{(1) Physical Nature.} On one hand, mouth shape, eye movement, and emotional expression all occur within the facial region and often exhibit mutual influence~\cite{tan2023emmn}, whereas head pose is independent of these components, characterized by rigid global rotation and translation. This independence makes pose disentanglement easier compared to disentangling movements within the facial region. On the other hand, in facial dynamics, mouth and eye movements occur more frequently than emotional expressions, making the extraction of mouth and eye features easier than that of emotional cues~\cite{wang2023progressive}. These physical characteristics motivate a hierarchical disentanglement strategy: we first disentangle head pose and facial movement (a relatively simpler task), then further separate mouth shape and eye movement, and finally disentangle emotional expression (a more complex task). \textbf{(2) Dataset Characteristics.} Existing datasets can generally be categorized into two types: those that feature neutral expressions with a wide range of poses and eye movements (e.g., HDTF~\cite{zhang2021flow}), and those that emphasize diverse expressions but involve minimal pose variations (e.g., MEAD~\cite{wang2020mead}). This dichotomy encourages us to utilize the former, where there is minimal interference from emotional expressions for disentangling pose, mouth, and eye movements, and the latter for learning disentangled emotional expressions. Therefore, we propose an efficient disentanglement strategy comprising Face-Pose Decoupling, Mouth-Eye Decoupling and Expression Decoupling, separating the overall facial dynamics into pose, mouth, eye and expression components.


\paragraph{Face-Pose Decouple}
As depicted at the top of Fig.~\ref{fig:overview} (b), we introduce cross-reconstruction technical, which involves synthesized images of switched faces: $I^{f_a}_{p_b}$ and $I^{f_b}_{p_a}$. Here, we superimpose the face region (\textit{i.e.,} mouth, eye and expression) of $I^a$ onto $I^b$ and vice versa. Subsequently, the encoder $E$ encodes them into canonical features, which are processed through $PLN$ and $FLN = MLN+EYLN+EXLN$ to obtain corresponding features:
\begin{equation}
    f^{p_b}, f^{f_a} = PLN(E(I^{f_a}_{p_b})), FLN(E(I^{f_a}_{p_b}))
\end{equation}
\begin{equation}
    f^{p_a}, f^{f_b} = PLN(E(I^{f_b}_{p_a})), FLN(E(I^{f_b}_{p_a}))
\end{equation}
Next, we substitute the extracted face features and feed them into the generator $G$ to perform cross reconstruction of the original images: $\hat{I}^b = G(f^{p_b}, f^{f_b})$ and $\hat{I}^a = G(f^{p_a}, f^{f_a})$. Additionally, we include identity features $f^{id}$ extracted from another frame of the same identity as input to the generator $G$. Afterward, we supervise the Face-Pose Decouple module by adopting reconstruction loss $\mathcal{L}_\text{rec}$, perceptual loss $\mathcal{L}_\text{per}$~\cite{johnson2016perceptual, zhang2018unreasonable} and adversarial loss $\mathcal{L}_\text{adv}$:
\begin{equation}
\label{eq:1}
   \mathcal{L}_\text{rec} = \sum_{\#={a,b}}\|I^\#-\hat{I}^\#\|_1; 
\end{equation}
\begin{equation}
   \mathcal{L}_\text{per} = \sum_{\#={a,b}}\|\Phi(I^\#)-\Phi(\hat{I}^\#)\|^2_2;
\end{equation}
\begin{equation}
\label{eq:2}
\mathcal{L}_\text{adv} = \sum_{\#={a,b}}(\text{log}D(I^\#)+\text{log}(1-D(\hat{I}^\#))),
\end{equation}
where $\Phi$ denotes the feature extractor of VGG19~\cite{simonyan2014very} and $D$ is a discriminator tasked with distinguishing between reconstructed images and ground truth (GT). In addition, self-reconstruction of the Ground Truth (GT) is crucial, where face features and pose features are extracted from the same image and then input into $G$ to reconstruct itself using $\mathcal{L}_\text{self}$. Furthermore, we impose feature-level constraints on the network:
\begin{equation}
\begin{aligned}
\label{eq:3}
\mathcal{L}_\text{fea} &= \sum_{\#={a,b}}(exp(-\mathcal{S}(f^{p_\#}, PLN(E(I^{\#})))) \\
&+ exp(-\mathcal{S}(f^{f_\#}, FLN(E(I^{\#}))))),
\end{aligned}
\end{equation}
where we extract mouth features and pose features from $I^a$ and $I^b$, aiming to minimize their disparity with those extracted from synthesized images of switched faces using cosine similarity $\mathcal{S}(\cdot,\cdot)$. Once the losses have converged, the parameters of $PLN$ are no longer updated for the remainder of training, significantly reducing training time and resource consumption for subsequent stages.

\paragraph{Mouth-Eye Decouple}
As shown in the middle of Fig.~\ref{fig:overview} (b), after successfully disentangling head pose from facial dynamics, we further decouple mouth shape and eye movement bn training $MLN$ and $EYLN$. Similar to the Face-Pose Decouple strategy, we swap the eye regions between two images and then reverse the eye features in the latent space to reconstruct the original images. However, it is worth noting that eye movements are subtler compared to head pose and mouth shape. Therefore, in addition to the loss function in last stage, we extract eye masks $m^{a,b}_{eye}$ and apply them to weight the reconstruction loss during training, emphasizing accurate reconstruction of the eye region:
\begin{equation}
\label{eq:2}
   \mathcal{L}_\text{rec} = \sum_{\#={a,b}}\|I^\#-\hat{I}^\#\|_1 \times(1+m^{\#}_{eye})
\end{equation}

Through this process, we successfully disentangle mouth shape and eye movement and then freeze the parameters of $MLN$ and $EYLN$, thereby reducing training time and computational cost.

\paragraph{Expression Decouple} As illustrated in the bottom of Fig.~\ref{fig:overview} (b), to decouple expression information from driving image $I^d$, we introduce Expression-aware Latent Navigation module ($EXLN$) and a lightweight plug-and-play Emotion Enhancement Module ($EEM$), both trained via self-reconstruction complementary learning. Specifically, given an identity source $I^i$ and a driving image $I^d$ sharing the same identity as $I^i$ but differing in mouth shapes, head poses, eye movements and emotional expressions, our pre-trained modules (i.e., $E$, $PLN$, $MLN$, $EYLN$,and $G$) from previous stage effectively disentangle head pose, mouth shape and eye movement from $I^d$ and drive $I^i$ to generate $\hat{I}^g_n$ with matching head pose, mouth shape and eye movement as $I^d$ but with the same expression with $I^i$. Therefore, to faithfully reconstruct $I^d$ with the same expression, $EXLN$ is compelled to learn complementary information not disentangled by $PLN$,$MLN,EYLN$, precisely the expression information. Motivated by the observation~\cite{wang2023progressive} that expression variation in a video sequence is typically less frequent than changes in other motions, we define a window of size $K$ around $I^d$ and average $K$ extracted expression features to obtain a clean expression feature $f^{r\rightarrow ex}$. $f^{r\rightarrow ex}$ is then combined with extracted pose, mouth and eye features as input to the generator $G$. Additionally, $EEM$ takes $f^{r\rightarrow ex}$ as input and utilizes affine transformations to produce $f^{ex} = (f^{ex}_s, f^{ex}_b)$ that control adaptive instance normalization (AdaIN)~\cite{huang2017arbitrary} operations. The AdaIN operations further adapt identity feature $f^{id}$ as emotion-conditioned features $f^{id}_{ex}$ by:
\begin{equation}
    f^{id}_{ex} := EEM(f^{id})= f^{ex}_s \frac{f^{id} - \mu(f^{id})}{\sigma(f^{id})} + f^{ex}_b,
\end{equation}
where $\mu(\cdot)$ and $\sigma(\cdot)$ represent the average and variance operations. Subsequently, we generate output $\hat{I}^g_{ex}$ with the expression of $I^d$ via Eq.~\ref{eq:g}. We enforce a motion reconstruction loss~\cite{wang2023progressive} $\mathcal{L}_\text{mot}$ in addition to the same reconstruction loss $\mathcal{L}_\text{rec}$, perceptual loss $\mathcal{L}_\text{per}$ and adversarial loss $\mathcal{L}_\text{adv}$ as Eq.~\ref{eq:1} and Eq.~\ref{eq:2}:
\begin{equation}
    \mathcal{L}_\text{mot} = \|\phi(I^d)-\phi(\hat{I}^g_{ex})\|_2 + \|\psi(I^d)-\psi(\hat{I}^g_{ex})\|_2,
\end{equation}
where $\phi(\cdot)$ and $\psi(\cdot)$ denote features extracted by the 3D face reconstruction network and the emotion network of~\cite{danvevcek2022emoca}. Moreover, to ensure that the synthesized image accurately mimics the mouth shape of the driving frame, we further introduce a mouth consistency loss $\mathcal{L}_\text{m-c}$:
\begin{equation}
    \mathcal{L}_\text{m-c} = e^{-\mathcal{S}(MLN(E(\hat{I}^g_{ex})), MLN(E(I^{d})))},
\end{equation}
where $MLN$ and $E$ are pretrained in the previous stage. During training, we only need to train lightweight $EXNL$ and $EEM$, resulting in fast training. 

\begin{figure}[t] 
  \includegraphics[width=\linewidth]{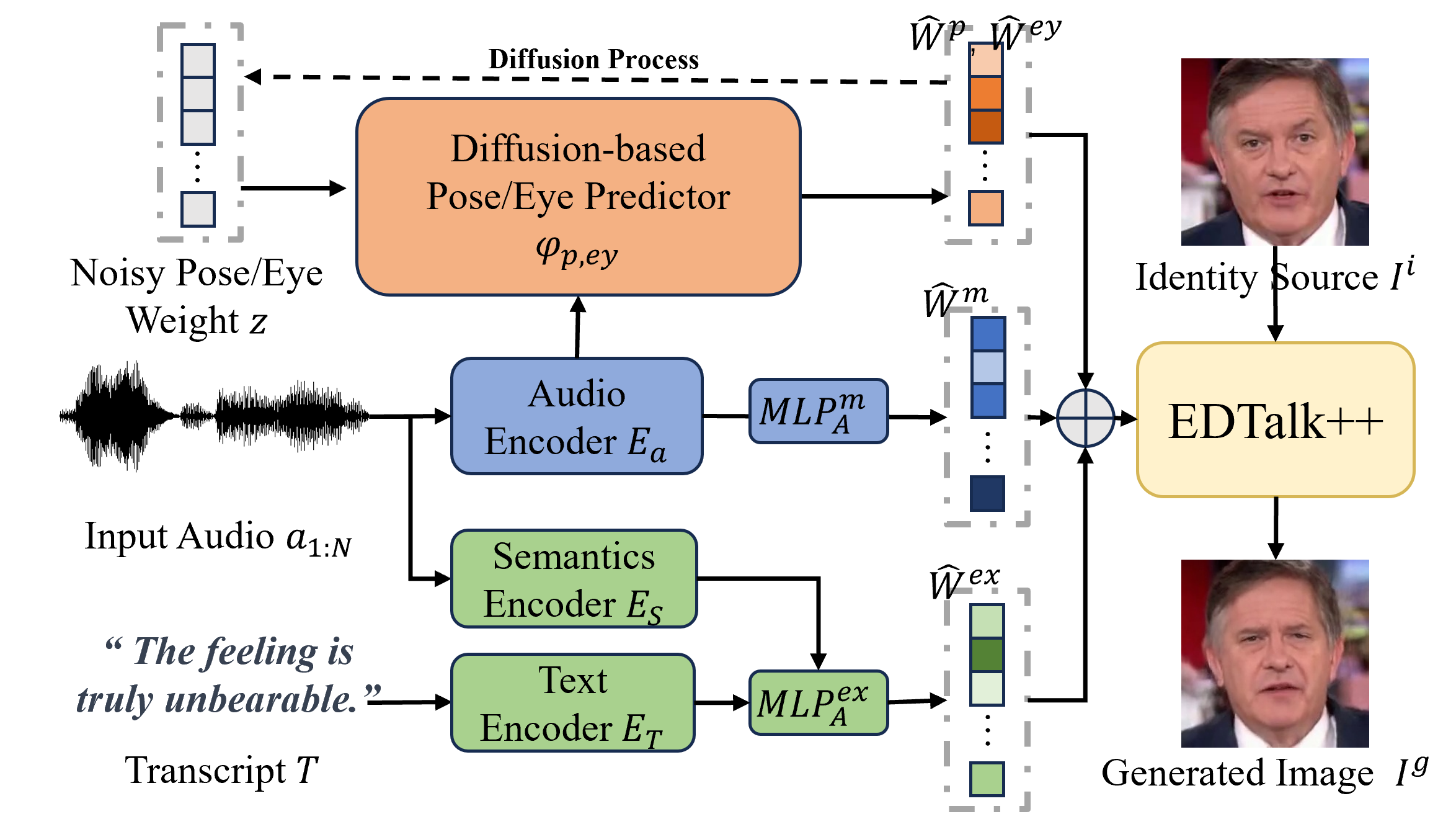}
    \caption{The overview of Audio-to-Motion, for mouth, pose, expression prediction.}
    \label{fig:audio_driven}
\end{figure}

After successfully training the three-stage Efficient Disentanglement module, we acquire four disentangled spaces, enabling one-shot video-driven talking face generation with separate control of identity, pose, mouth shape, eye movement and expression, given different driving sources, as illustrated in Fig.~\ref{fig:overview} (a).

\subsection{Audio-to-Motion}
\label{sec:audio_driven}

Integrating the disentangled spaces, we aim to address a more appealing but challenging task: audio-driven talking face generation. In this section, depicted in Fig.~\ref{fig:audio_driven}, we introduce three modules to predict the weights of pose\&eye, mouth, and expression from audio. These modules replace the driving video input, facilitating audio-driven talking face generation.

\paragraph{Audio-Driven Lip Generation}
Prior works~\cite{ma2023styletalk, tan2023emmn} generate facial dynamics, encompassing lip motions and expressions, in a holistic manner, which proves challenging for two main reasons: 1) Expressions, being acoustic-irrelevant motions, can impede lip synchronization~\cite{zhang2023sadtalker}. 2) The absence of lip visual information hinders fine details synthesis at the phoneme level~\cite{park2022synctalkface}. Thanks to the disentangled mouth space obtained in the previous stage, we naturally mitigate the influence of expression without necessitating special training strategies or loss functions like~\cite{zhang2023sadtalker}. Additionally, since the decoupled space is trained during video-driven talking face generation using video as input, which offers ample visual information in the form of mouth bases $b^m_i$ stored in the bank $B^m$, we eliminate the need for extra visual memory like~\cite{park2022synctalkface}. Instead, we only need to predict the weight $w^m_i$ of each base $b^m_i$, which generates the fine-grained lip motion. To achieve this, we design an Audio Encoder $E_a$, which embeds the audio feature into a latent space $f^a = E_a(a_{1:N})$. Subsequently, a linear layer $MLP^m_A$ is added to decode the mouth weight $\hat{W}^m$. During training, we fix the weights of all modules and only update $E_a$ and $MLP^m_A$ using the weighted sum of feature loss $\mathcal{L}^m_{fea}$, reconstruction loss $\mathcal{L}^m_{rec}$ and sync loss $\mathcal{L}^m_{sync}$~\cite{prajwal2020lip}:
\begin{equation}
    \mathcal{L}^m_{fea} = \|W^m - \hat{W}^m\|_2,\qquad
    \mathcal{L}^m_{rec} = \|I - \hat{I}\|_2,
\end{equation}
\begin{equation}
\mathcal{L}^m_{sync} = -\text{log}(\frac{v\cdot s}{max(\|v\|_2 \cdot \|s\|_2, \epsilon)}),
\end{equation}
where $W^m = MLN(E(I))$ is the GT mouth weight extracted from GT image $I$ and $\hat{I}$ is generated image using Eq.~\ref{eq:g}. $\mathcal{L}^m_{sync}$ is introduced from~\cite{prajwal2020lip}, where $v$ and $s$ are extracted by the speech encoder and image encoder in SyncNet~\cite{chung2017out}.


\paragraph{Diffusion-Based Probabilistic Pose\&Eye Generation}
Due to the nature of one-to-many mapping from the input audio to head poses and eye movements, learning a deterministic mapping like previous works~\cite{wang2021audio2head,wang2022one, zhou2020makelttalk} output the same results, which bring ambiguity and inferior visual results. To generate probabilistic and realistic head motions and eye movements, we predict the pose and eye weights $\hat{W}^p,\hat{W}^{ey}$ using diffusion model $\varphi_{p,ey}$~\cite{DDIM,DDPM}, as illustrated in Fig.~\ref{fig:audio_driven}. During training (indicated by dashed lines), we extract pose and eye weights $W^{p,ey}$ from videos as the ground truth and feed them into the diffusion process to obtain the noised weights $z$. During the denoising process, we train a Diffusion-based Pose/Eye Predictor $\varphi_{p,ey}$ to generate pose and eye weights $\hat{W}^{p,ey} = \varphi_{p,ey}(z, f^a_t)$. This process ensures the diversity of head motions and eye movements while maintaining consistency with the audio rhythm.

\paragraph{Semantically-Aware Expression Generation}
As finding videos with a desired expression may not always be feasible, potentially limiting their application~\cite{ma2023talkclip}, we aim to explore the emotion contained in audio and transcript with the aid of the introduced Semantics Encoder $E_S$ and Text Encoder $E_T$. Inspired by~\cite{wang2021fine}, our Semantics Encoder $E_S$ is constructed upon the pretrained HuBERT model~\cite{hsu2021hubert}, which consists of a CNN-based feature encoder and a transformer-based encoder. We freeze the CNN-based feature encoder and only fine-tuned the transformer blocks. Text Encoder $E_T$ is inherited from the pretrained Emoberta~\cite{kim2021emoberta}, which encodes the overarching emotional context embedded within textual descriptions. We concatenate the embeddings generated by $E_S$ and $E_T$ and feed them into a $MLP^{ex}_A$ to generate the expression weights $\hat{W}^{ex}$. Since audio or text may not inherently contain emotion during inference, such as in TTS-generated speech, in order to support the prediction of emotion from a single modality, we randomly mask ($\mathcal{M}$) a modality with probability $p$ during training, inspired by HuBERT: 
\begin{equation}
\label{eq6}
\hat{W}^{ex}=\left\{
\begin{aligned}
MLP^{ex}_a(E_S(a), E_T(T)) & , & 0.5 \le p \le 1, \\
MLP^{ex}_a(\mathcal{M}(E_S(a)), E_T(T)) & , & 0.25 \le p < 0.5, \\
MLP^{ex}_a(E_S(a), \mathcal{M}(E_T(T))) & , & 0 \le p < 0.25. \\
\end{aligned}
\right.
\end{equation}

We employ $\mathcal{L}_\text{exp} = \|W^{ex} - \hat{W}^{ex}\|_1$ to encourage $\hat{W}^{ex}$ close to weight $W^{ex}$ generated by pretrained $EXLN$ from emotional frames. Until now, we are able to generate \textbf{probabilistic} \textbf{semantically-aware} talking head videos solely from an identity image and the driving audio.

\begin{figure*}[t]
  \centering
  \includegraphics[width=0.95\linewidth]{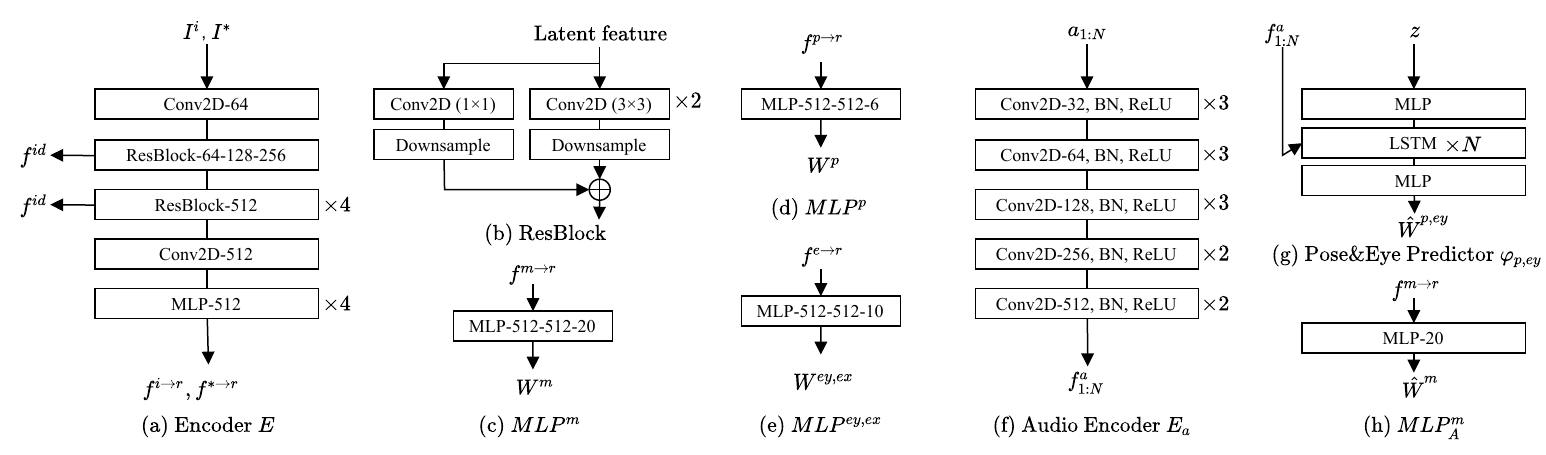}
    \caption{Detailed architecture for different components in our EDTalk++.}
    \label{fig:detail}
    \vspace{-0.2in}
\end{figure*}

\subsection{Network Architecture}

We extend the generator architecture of LIA~\cite{wang2021latent} from a resolution of $256 \times 256$ to $512 \times 512$ by introducing additional convolutional layers. Here, we delineate the details of the other network architectures depicted in Fig.~\ref{fig:detail}.

\paragraph{Encoder $E$} The component projects the identity source $I^i$ and driving source $I^*$ into the identity feature $f^{id}$ and the latent features $f^{i \rightarrow r}$, $f^{* \rightarrow r}$. It comprises several convolutional neural networks (CNN) and ResBlocks. The outputs of ResBlock serve as the identity feature $f^{id}$, which is then fed into Generator $G$ to enrich identity information through skip connections. Subsequently, four multi-layer perceptrons (MLP) are employed to generate the latent features $f^{i \rightarrow r}$, $f^{* \rightarrow r}$.

\paragraph{$MLP^m$, $MLP^p$, $MLP^{ey}$, $MLP^{ex}$ and $MLP^m_A$} To achieve efficient training and inference, these four modules are implemented with five simple MLPs.

\paragraph{Audio Encoder $E_a$} This network takes audio feature sequences $a_{1:T}$ as input. These sequences are passed through a series of convolutional layers to produce audio feature $f^a_{1:N}$.

\paragraph{Pose\&Eye Predictor $\varphi_{p,ey}$} $\varphi_{p,ey}$ comprises MLP layer, $N$ LSTM layer and the output MLP layer.

\begin{table*}[t]
	\centering
	\resizebox{\linewidth}{!}{
	\begin{tabular}{l|cccccc|ccccc}
		\toprule
		\multicolumn{1}{c}{\multirow{2}[4]{*}{\textbf{Method}}} & \multicolumn{6}{c}{\textbf{MEAD}~\cite{wang2020mead}} & \multicolumn{5}{c}{\textbf{HDTF}~\cite{zhang2021flow}}\\
		\cmidrule(lr){2-7}  \cmidrule(lr){8-12}  \multicolumn{1}{c}{} & \multicolumn{1}{c}{PSNR$\uparrow$}& \multicolumn{1}{c}{SSIM$\uparrow$} & \multicolumn{1}{c}{M/F-LMD$\downarrow$} & \multicolumn{1}{c}{FID$\downarrow$}  & \multicolumn{1}{c}{$\text{Sync}_\text{conf}\uparrow$}& \multicolumn{1}{c}{$\text{Acc}_\text{emo}\uparrow$}& \multicolumn{1}{c}{PSNR$\uparrow$} & \multicolumn{1}{c}{SSIM$\uparrow$} & \multicolumn{1}{c}{M/F-LMD$\downarrow$} & \multicolumn{1}{c}{FID$\downarrow$} & \multicolumn{1}{c}{$\text{Sync}_\text{conf}\uparrow$}
		\\

		\midrule
		MakeItTalk~\cite{zhou2020makelttalk} & 19.442 & 0.614 & 2.541/2.309 & 37.917 & 5.176& 14.64 & 21.985 & 0.709 & 2.395/2.182 & 18.730 & 4.753    \\
		Wav2Lip~\cite{prajwal2020lip} & 19.875 & 0.633 & 1.438/2.138 & 44.510 & \textbf{8.774}& 13.69 & 22.323 & 0.727 & 1.759/2.002 & 22.397 & \textbf{9.032}   \\
		Audio2Head~\cite{wang2021audio2head} & 18.764 & 0.586 & 2.053/2.293 & 27.236 & 6.494& 16.35 & 21.608 & 0.702 & 1.983/2.060 & 29.385 & 7.076   \\
		PC-AVS~\cite{zhou2021pose} & 16.120 & 0.458 & 2.649/4.350 & 38.679 & 7.337&12.12& 22.995 & 0.705 & 2.019/1.785 & 26.042 & 8.482  \\

		AVCT~\cite{wang2022one} & 17.848 & 0.556 & 2.870/3.160 & 37.248 & 4.895& 13.13 & 20.484 & 0.663 & 2.360/2.679 & 19.066 & 5.661 \\
		SadTalker~\cite{zhang2023sadtalker} & 19.042 & 0.606 & 2.038/2.335 & 39.308 & 7.065& 14.25 & 21.701 & 0.702 & 1.995/2.147 & 14.261 & 7.414  \\
		IP-LAP~\cite{zhong2023identity} & 19.832 & 0.627 & 2.140/2.116 & 46.502 & 4.156& 17.34 & 22.615 & 0.731 & 1.951/1.938 & 19.281 & 3.456  \\
        TalkLip~\cite{wang2023seeing} & 19.492 & 0.623 & 1.951/2.204 & 41.066 & 5.724& 14.00 & 22.241 & 0.730 & 1.976/1.937 & 23.850 & 1.076\\
		\midrule
		EAMM~\cite{ji2022eamm} & 18.867 & 0.610 & 2.543/2.413 & 31.268 & 1.762& 31.08 & 19.866 & 0.626 & 2.910/2.937 & 41.200 & 4.445   \\
		StyleTalk~\cite{ma2023styletalk} &21.601 & 0.714 & 1.800/1.422 & 24.774 & 3.553& 63.49 & 21.319 & 0.692 & 2.324/2.330 & 17.053 & 2.629  \\
		PD-FGC~\cite{wang2023progressive} & 21.520 & 0.686 & 1.571/1.318 & 30.240 & 6.239& 44.86 & 23.142 & 0.710 & \textbf{1.626}/1.497 & 25.340 & 7.171   \\
		EAT~\cite{gan2023efficient} & 20.007 & 0.652 & 1.750/1.668 & 21.465 & 7.984 & 64.40& 22.076 & 0.719 & 2.176/1.781 & 28.759 & 7.493   \\ 
		EDTalk & {21.628} & {0.722} & {1.537}/{1.290} & {17.698} & 8.115& {67.32} &{25.156} & {0.811} & 1.676/{1.315} & {13.785} & 7.642   \\ 
        EDTalk++ & \textbf{22.423} & \textbf{0.756} & \textbf{1.213}/\textbf{1.1960} & \textbf{15.679} &8.036&\textbf{68.21} & \textbf{25.914} & \textbf{0.824} & \textbf{1.277}/\textbf{1.230} & \textbf{13.224} & 7.679\\
		\midrule
		GT & 1.000 & 1.000     &0.000/0.000   &   0.000  & 7.364 & 79.65   &   1.000  & 1.000     &0.000/0.000   &   0.000     & 7.721  \\
		\bottomrule
	
	\end{tabular}%
	}
	\caption{Quantitative comparisons with state-of-the-art methods.}
	\label{tab:quantitative}%
\end{table*}%

\begin{table*}[t]
	\centering
	\resizebox{\linewidth}{!}{
	\begin{tabular}{l|ccccc|ccccc}
		\toprule
		\multicolumn{1}{c}{\multirow{2}[4]{*}{\textbf{Method}}} & \multicolumn{5}{c}{\textbf{Voxceleb2}~\cite{chung2018voxceleb2}} & \multicolumn{5}{c}{\textbf{LRW}~\cite{chung2017lip}}\\
		\cmidrule(lr){2-6}  \cmidrule(lr){7-11}  \multicolumn{1}{c}{} & \multicolumn{1}{c}{PSNR$\uparrow$}& \multicolumn{1}{c}{SSIM$\uparrow$} & \multicolumn{1}{c}{M-LMD$\downarrow$} & \multicolumn{1}{c}{F-LMD$\downarrow$}  & \multicolumn{1}{c}{$\text{Sync}_\text{conf}\uparrow$}& \multicolumn{1}{c}{PSNR$\uparrow$} & \multicolumn{1}{c}{SSIM$\uparrow$} & \multicolumn{1}{c}{M-LMD$\downarrow$} & \multicolumn{1}{c}{F-LMD$\downarrow$} & \multicolumn{1}{c}{$\text{Sync}_\text{conf}\uparrow$}
		\\

		\midrule
		MakeItTalk~\cite{zhou2020makelttalk} & 20.526 & 0.706 & 2.435 & 2.380 & 3.896 & 22.334 & 0.729 & 2.099 & 1.960 & 3.137    \\
		Wav2Lip~\cite{prajwal2020lip} & 20.760 & 0.723 & 2.143 & 2.182 & \textbf{8.680} & 23.299 & 0.764 & 1.699 & 1.703 &\textbf{ 7.545}    \\
		Audio2Head~\cite{wang2021audio2head} & 17.344 & 0.577 & 3.651 & 3.712 & 5.541 & 18.703 & 0.601 & 2.866 & 3.435 & 5.428  \\
		PC-AVS~\cite{zhou2021pose} & 21.643 & 0.720 & 2.088 & 1.830 & 7.928 & 16.744 & 0.509 & 5.603 & 4.691 & 3.622  \\

		AVCT~\cite{wang2022one} & 18.751 & 0.645 & 2.739 & 3.062 & 4.238 & 21.188 & 0.689 & 2.290 & 2.395 & 3.927  \\
		SadTalker~\cite{zhang2023sadtalker} & 20.278 & 0.700 & 2.252 & 2.388 & 6.356 & - & - & - & - & -  \\
		IP-LAP~\cite{zhang2023sadtalker} & 20.955 & 0.724 & 2.125 & 2.154 & 3.295 & {23.727} & 0.770 & 1.779 & 1.683 & 3.027  \\
        TalkLip~\cite{wang2023seeing} & 20.633 & 0.723 & 2.084 & 2.191 & 6.520 & 22.706 & 0.751 & 1.803 & 1.770 & 6.021 \\
		\midrule
		EAMM~\cite{ji2022eamm} &  17.038 & 0.562 & 4.172 & 4.163 & 3.815 & 18.643 & 0.607 & 3.593 & 3.773 & 3.414   \\
		StyleTalk~\cite{ma2023styletalk} &21.112 & 0.722 & 2.113 & 2.136 & 2.120 & 21.283 & 0.705 & 2.394 & 2.142 & 2.430  \\
		PD-FGC~\cite{wang2023progressive} & 22.110 & 0.729 & \textbf{1.743} & 1.630 & 6.686 & 22.481 & 0.711 & {1.576} & 1.534 & 6.119    \\ 
		EAT~\cite{gan2023efficient} & 20.370 & 0.689 & 2.586 & 2.383 & 6.864 & 21.384 & 0.704 & 2.128 & 1.927 & 6.630   \\ 
		EDTalk & {22.107} & {0.763} & 1.851 & {1.608} & 6.591 & 23.409 & {0.779} & 1.729 & {1.379} & 6.914    \\ 
        EDTalk++ & \textbf{22.117} & \textbf{0.775} & 1.831 & \textbf{1.531} & 6.675 & \textbf{23.977} & \textbf{0.802} & \textbf{1.310} & \textbf{1.201} & 6.713\\
		\midrule
		GT & 1.000 & 1.000     &0.000/0.000   &   0.000  & 6.808 &   1.000  & 1.000     &0.000/0.000   &   0.000     & 6.952 \\
		\bottomrule
	
	\end{tabular}%
	}
	\caption{\textbf{Quantitative comparisons with state-of-the-art methods.} We test each method on Voxceleb2 and LRW datasets, and the best scores in each metric are highlighted in bold. The symbol $"\uparrow"$ and  $"\downarrow"$ indicate higher and lower metric values for better results, respectively.}
	\label{tab:qauantitative_supp}%
    \vspace{-0.2in}
\end{table*}%

\section{Experiments}

We verify the effectiveness of our proposed approach through comprehensive experiments conducted on multiple publicly accessible benchmarking datasets for talking head generation. In this section, we will illustrate the experimental setup first and then analyze our experimental results.

\subsection{Dataset Description}

Our model is trained and evaluated on the datasets MEAD~\cite{wang2020mead}, HDTF~\cite{zhang2021flow} and VFHQ~\cite{vfhq}. Additionally, we report results on additional datasets, including  LRW~\cite{chung2017lip} and Voxceleb2~\cite{chung2018voxceleb2}, for further assessment of our method. 

\paragraph{MEAD} MEAD entails 60 speakers, with 43 speakers accessible, delivering 30 sentences expressing eight emotions at three varying intensity levels in a laboratory setting. Consistent with prior studies~\cite{ji2022eamm, gan2023efficient}, we designate videos featuring speakers identified as `M003,' `M030,' `W009,' and `W015' for testing, while the videos of the remaining speakers are allocated for training.

\paragraph{HDTF} The videos of the HDTF dataset are collected from YouTube, renowned for their high quality, high definition content, featuring over 300 distinct identities. To facilitate training and testing, we partition the dataset using an 8:2 ratio based on speaker identities, allocating 80\% for training and 20\% for testing.

\paragraph{VFHQ} VFHQ contains over 16,000 high-quality clips from various interview scenarios, featuring a wide range of head poses and eye movements.

\paragraph{Voxceleb2} Voxceleb2~\cite{chung2018voxceleb2} is a large-scale talking head dataset, boasting over 1 million utterances from 6,112 celebrities. It's important to note that we solely utilize Voxceleb2 for evaluation purposes, selecting 200 videos randomly from its extensive collection.

\paragraph{LRW} LRW~\cite{chung2017lip} is a word-level dataset comprising more than 1000 utterances encompassing 500 distinct words. For evaluation, we randomly select 500 videos from the dataset.

For video preprocessing, we employ face cropping and resize the cropped videos to the resolution of $512 \times 512$ for training and testing following FOMM~\cite{siarohin2019first}. Adhere to Wav2Lip~\cite{prajwal2020lip}, audio is down-sampled to 16 kHz and transformed into mel-spectrograms using an FFT window size of 800, hop length of 200, and 80 Mel filter banks. During the evaluation, for datasets without emotional labels, we utilize the first frame of each video as the source image and the corresponding audio as the driving audio to generate talking head videos. For emotional videos sourced from MEAD, we use the video itself as an expression reference. We select a frame with a `Neutral' emotion from the same speaker as the source image for emotional talking head synthesis.

\begin{figure*}[t]
  \centering
  \includegraphics[width=0.9\linewidth]{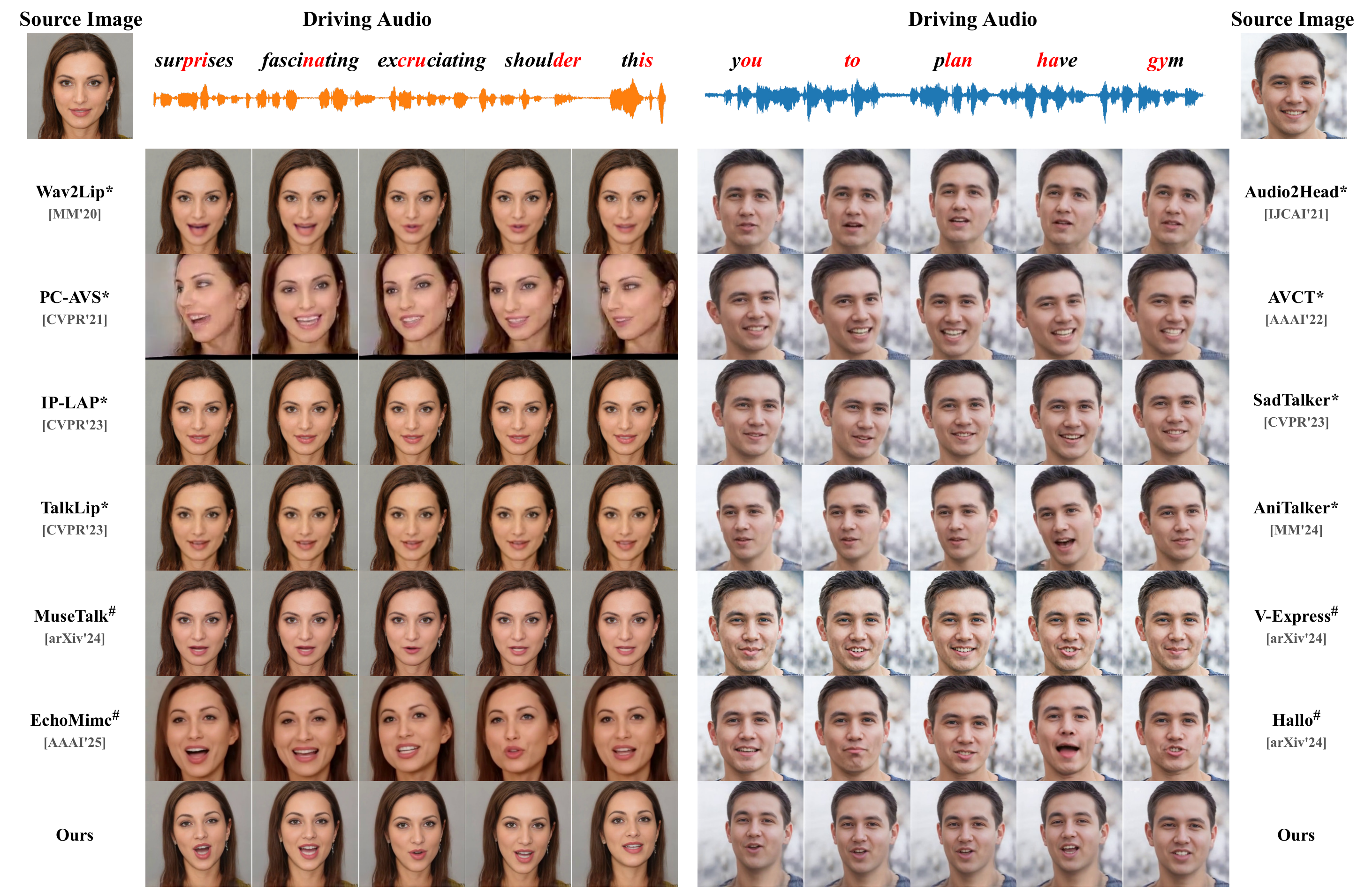}
    \caption{Qualitative comparisons with state-of-the-art emotion-agnostic talking head generation methods.}
    \vspace{-0.2in}
    \label{fig:comparison_neu}
\end{figure*}

\begin{figure*}[t]
  \centering
  \includegraphics[width=0.9\linewidth]{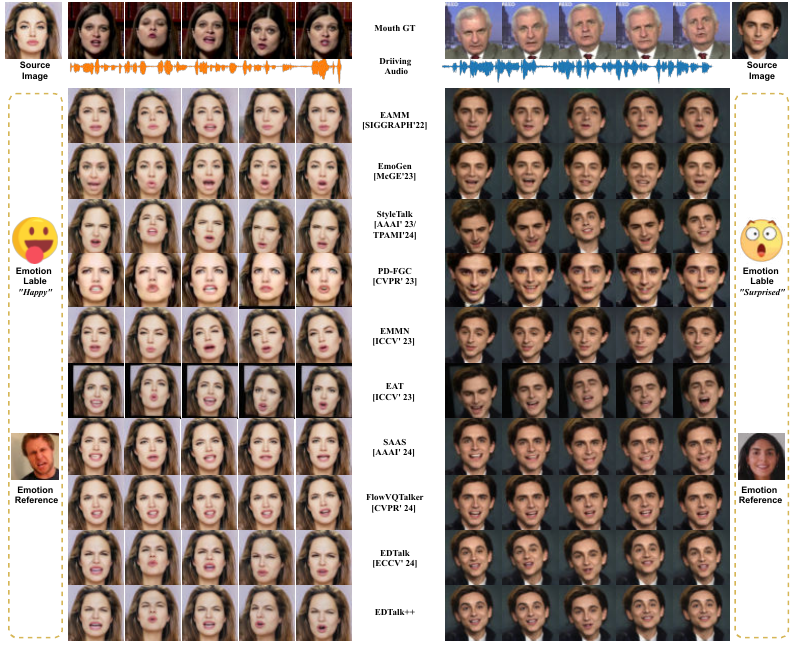}
    \caption{Qualitative comparisons with state-of-the-art emotional talking head generation methods.}
    \vspace{-0.2in}
    \label{fig:comparison_emo}
\end{figure*}

\subsection{Implementation Details}
Our method is implemented using PyTorch and trained using the Adam optimizer. The dimension of the latent code $f^{*\rightarrow r}$ and bases $b^*$ is set to 512, and the number of bases of $B^m$, $B^p$, $B^{ey}$ and $B^{ex}$ are set to 20, 6, 10 and 10, respectively. The weight for $\mathcal{L}_\text{mot}$ is set to 10 and the remaining weights are set to 1. The encoder $E$ and generator $G$ are pre-trained in a similar setting as LIA~\cite{wang2021latent}. Subsequently, we freeze the weights of the encoder $E$ and generator $G$, focusing solely on training the Face-Pose Decouple Module and Mouth-Eye Decouple Module. In this stage, our model is trained exclusively on the emotion-agnostic HDTF and VFHQ dataset, where videos consistently exhibit a `Neutral' emotion alongside diverse head poses and eye movements. It ensures that the modules concentrates solely on variations in head pose, mouth shape and eye movement, avoiding the encoding of expression-related information. All loss function weights are set to 1. The training process typically requires approximately one hour, employing a batch size of 4 and a learning rate of 2e-3, executed on 4 NVIDIA GeForce GTX 3090 GPUs with 24GB memory. Once these two modules are trained, we freeze all trained parameters and solely update the expression-related modules, including $MLP^{ex}$, expression bases $B^{ex}$, and the Expression Enhance Module $EEM$, utilizing MEAD and VFHQ datasets. We train our Audio-to-Lip model on the HDTF dataset for 30k iterations with a batch size of 4, requiring approximately 7 hours of computation on 4 NVIDIA GeForce GTX 3090 GPUs with 24GB memory. The Audio-to-Pose\&Eye model is trained on the VFHQ dataset for four hours.


\paragraph{Comparison Setting}
We compare our method with: (a) emotion-agnostic talking face generation methods: MakeItTalk~\cite{zhou2020makelttalk}, Wav2Lip~\cite{prajwal2020lip}, Audio2Head~\cite{wang2021audio2head}, PC-AVS~\cite{zhou2021pose}, AVCT~\cite{wang2022one}, SadTalker~\cite{zhang2023sadtalker}, IP-LAP~\cite{zhong2023identity}, TalkLip~\cite{wang2023seeing}, Anitalker~\cite{liu2024anitalker}, MuseTalk$^\#$~\cite{musetalk}, V-Express$^\#$~\cite{VExpress}, EchoMimic$^\#$~\cite{echomimic} and Hallo$^\#$~\cite{xu2024hallo}, where $^\#$ refers to diffusion-based models, which require more datasets and computation resource. (b) Emotional talking face generation methods: EAMM~\cite{ji2022eamm}, StyleTalk~\cite{ma2023styletalk}, PD-FGC~\cite{wang2023progressive}, EMMN~\cite{tan2023emmn}, EAT~\cite{gan2023efficient}, EmoGen~\cite{goyal2023emotionally}, SAAS~\cite{tan2024say}, FlowVQTalker~\cite{tan2024flowvqtalker}. Different from previous work, EDTalk++ encapsulates the entire face generation process without any other sources (e.g. poses~\cite{ji2022eamm, gan2023efficient}, 3DMM~\cite{ma2023styletalk, yin2022styleheat}, phoneme~\cite{ma2023styletalk, wang2022one}) and pre-processing operations during inference, which facilitates the application. We evaluate our model \textit{w.r.t.} (i) generated video quality using PSNR, SSIM~\cite{ZhouWang2004ImageQA} and FID~\cite{Seitzer2020FID}. (ii) audio-visual synchronization using Landmarks Distances on the Mouth (M-LMD)~\cite{chen2019hierarchical} and the confidence score of SyncNet~\cite{chung2017out}. (ii) emotional accuracy using $\text{Acc}_\text{emo}$ calculated by pretrained Emotion-Fan~\cite{meng2019frame} and Landmarks Distances on the Face (F-LMD).

\subsection{Experimental Results}

\paragraph{Quantitative Results} The quantitative results are presented in Tab.~\ref{tab:quantitative}, where our EDTalk++ achieve the best performance across most metrics, except $\text{Sync}_\text{conf}$. Wav2Lip pretrains their SyncNet discriminator on a large dataset~\cite{afouras2018deep}, which might lead the model to prioritize achieving a higher $\text{Sync}_\text{conf}$ over optimizing visual performance. It is evident in the blur mouths generated by Wav2Lip and inferior M-LMD score to our method. 

Apart from the quantitative assessments conducted on the MEAD and HDTF datasets, we present additional quantitative comparisons on Voxceleb2~\cite{chung2018voxceleb2} and LRW~\cite{chung2017lip}. The comparison results outlined in Tab.~\ref{tab:qauantitative_supp} demonstrate that our method outperforms state-of-the-art approaches across various metrics. IP-LAP~\cite{zhong2023identity} merely alters the mouth shape of the source image while maintaining the same head pose and expression, hence achieving a higher PSNR score. PD-FGC~\cite{wang2023progressive} attains superior M-LMD performance by training on Voxceleb2, a dataset comprising over \textbf{1 million} utterances from \textbf{6,112} celebrities, totaling \textbf{2400 hours} of data, which is \textbf{hundreds of times} larger than our dataset. Nevertheless, we still outperform PD-FGC in terms of F-LMD. SadTalker~\cite{zhang2023sadtalker} encounters challenges in processing even one second of audio, leading to the failure to generate talking face videos on the LRW dataset, where all videos are one second in duration.

\begin{figure}[t] 
  \includegraphics[width=\linewidth]{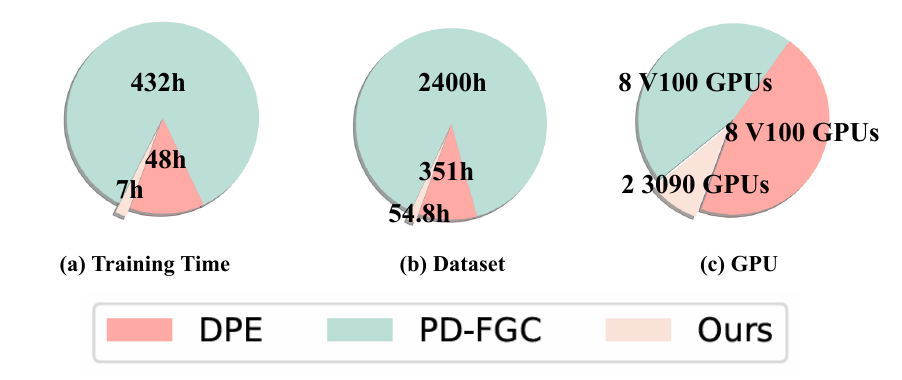} 
  \caption{Resources for training.}
  \label{fig:time}
  \vspace{-0.2in}
\end{figure}

\begin{table}[t]
  \resizebox{\linewidth}{!}{

  \begin{tabular}{@{}l|cccccc@{}}
    \toprule
    Metric/Method & TalkLip  & IP-LAP  & EAMM & EAT &EDTalk++ &GT \\
    \midrule
    Lip-sync & 3.31 &3.42  &3.49&3.85&\textbf{4.13}&4.74 \\
    Realness & 3.14 & 3.13  &3.26 &3.75&\textbf{4.92}& 4.81 \\
    $\text{Acc}_\text{emo}$ (\%) & 19.7 & 17.6  &44.3 &59.7&\textbf{64.5}& 75.6 \\
    \bottomrule
  \end{tabular}
  }
  \caption{User study results.}
  \label{tab:user_study}
  \vspace{-0.2in}
\end{table}
\paragraph{Qualitative Results}
Fig.~\ref{fig:comparison_neu} demonstrates visual results compared to emotion-agnostic talking face generation methods. Wav2Lip, IP-LAP, TalkLip, and MuseTalk only modify the mouth region while keeping the rest of the source image unchanged, resulting in unrealistic outputs. AVCT and SadTalker directly predict head poses from audio, but are also affected by inter-frame jitter. Audio2Head, PC-AVS, V-Express, and EchoMimic suffer from identity degradation. Additionally, Hallo exhibits artifacts in some subsequent frames over time. Compared to prior methods, our EDTalk++ effectively preserves identity and generates natural head poses, while maintaining visual consistency without introducing artifacts even in long-term video synthesis.

Fig.~\ref{fig:comparison_emo} presents a comparative analysis between our method and existing emotional talking head generation approaches. EmoGen, EAMM, and PD-FGC suffer from identity loss, which undermines the realism of the generated results. Besides, EmoGen and EAMM fail to accurately reproduce the target expressions. Due to its reliance on discrete emotion inputs, EAT is unable to synthesize fine-grained expressions. For example, it cannot replicate the subtle eye narrowing observed in the expression reference. For the ``happy'' expression, EAT exhibits unintended closed eyes, while PD-FGC generates unnatural teeth shapes. In contrast, EDTalk++ demonstrates superior performance in rendering realistic emotional expressions, achieving precise lip synchronization, and maintaining accurate head poses throughout the video.

\begin{figure}[t] 
  \includegraphics[width=\linewidth]{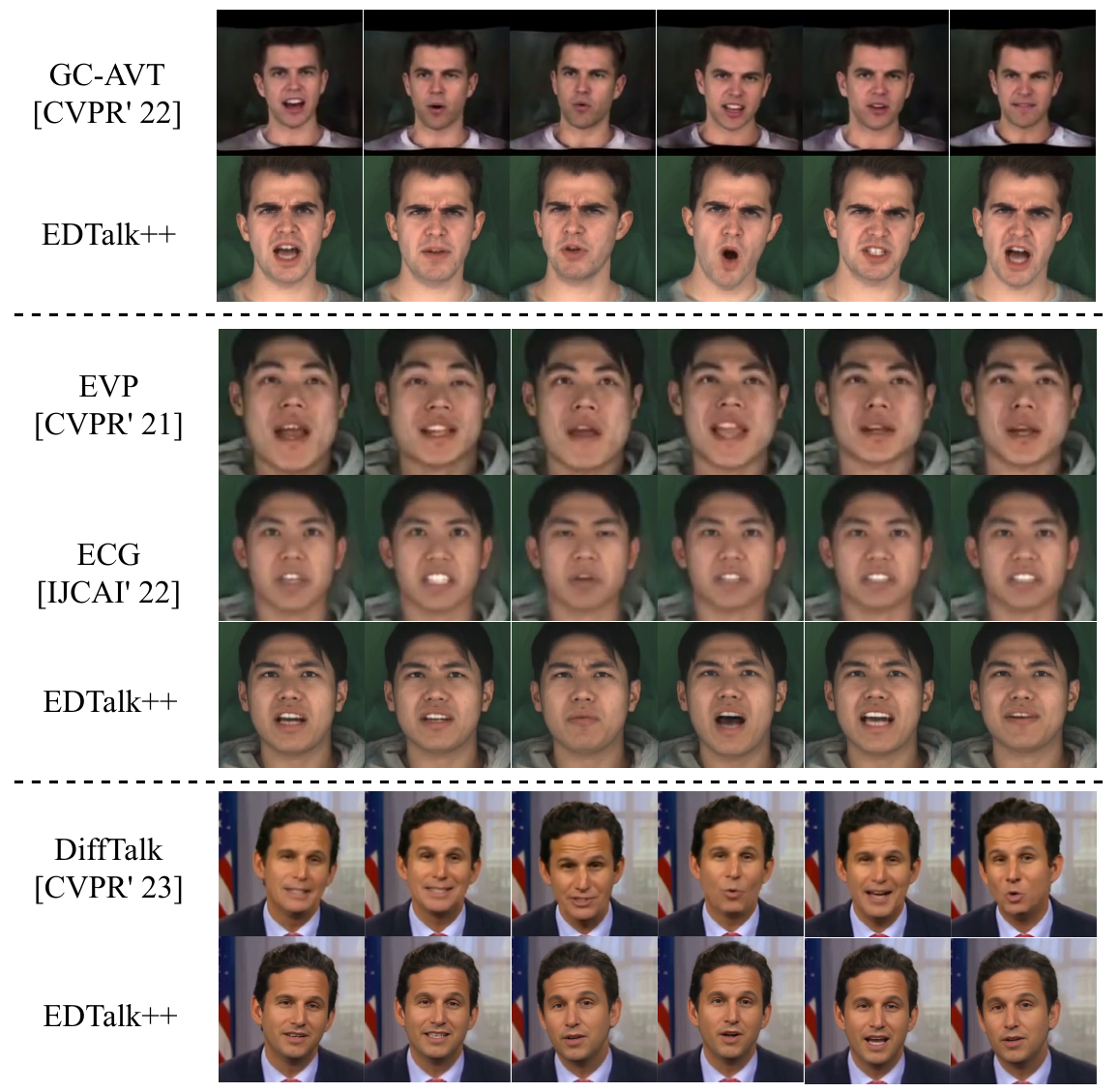} 
  \caption{Comparison results with SOTA methods that have not released their codes and pretrained models.}
  \vspace{-0.2in}
  \label{fig:more}
\end{figure}

We further conduct the comparison experiments with several SOTA talking face generation methods, including: GC-AVT~\cite{liang2022expressive}, EVP~\cite{ji2021audio}, ECG~\cite{sinhaemotion} and DiffTalk~\cite{shen2023difftalk}. However, due to the unavailability of codes and pre-trained models for these methods (except EVP), we can only extract video clips from the provided demo videos for comparison. The results are demonstrated in Fig.~\ref{fig:more}. Specifically, EVP and ECG are emotional talking face generation methods that utilize one-hot labels for emotional guidance, with EVP being a person-specific model and ECG being a one-shot method. Our method outperforms these methods in terms of emotional expression, while the teeth generated by ECG contribute to slightly unrealistic results. GC-AVT aims to mimic emotional expressions and generate accurate lip motions synchronized with input speech, resembling the setting of our EDTalk++. However, compared to EDTalk++, GC-AVT struggles to preserve reference identity, resulting in significant identity loss. DiffTalk is hindered by severe mouth jitter, which is more evident in the Supplementary Video (4:46-5:05).


\paragraph{User Study} We conduct a user study to evaluate our method for \textit{human likeness} test. We generate 10 videos for each method and invite 20 participants (10 males, 10 females) to score from 1 (worst) to 5 (best) in terms of lip-synchronization, realness, and emotion classification. The average scores reported in Tab.~\ref{tab:user_study} demonstrate that our method achieves the best performance in all aspects.

\paragraph{Efficiency analysis}
Our approach is highly efficient in terms of training time, required data and computational resources in decoupling spaces. Considering that PD-FGC~\cite{wang2023progressive} conducts experiments at a resolution of 224×224 and DPE operates at 256×256, and that DPE~\cite{pang2023dpe} is only capable of decoupling expressions from head poses, we ensure a fair comparison by conducting our experiments at the same resolution of 256×256. Additionally, to align with DPE's settings, we omit the eye decoupling component in our pipeline. In other words, the disentanglement process is simplified to the mouth-pose decoupling stage and the expression decoupling stage. In the mouth-pose decoupling stage, we solely utilize the HDTF dataset, containing \textbf{15.8 hours} of videos, for the decoupling. Training with a batch size of 4 on \textbf{two 3090 GPUs} for \textbf{4k} iterations achieves state-of-the-art performance, which takes about one hour. In contrast, DPE is trained on the VoxCeleb dataset, which comprises \textbf{351 hours} of video, for \textbf{100K} iterations initially, then an additional \textbf{50K} iterations with a batch size of 32 on 8 V100 GPUs, which takes over 2 days. Besides, they need to train two task-specific generators for expression and pose. Similarly, PD-FGC takes \textbf{2 days} on \textbf{4 Tesla V100 GPUs} for lip, and another \textbf{2 days on 4 Tesla V100 GPUs} for pose decoupling. It significantly exceeds our computational resources and training time. In the expression decouple stage, we train our model on MEAD and HDTF dataset (total \textbf{54.8 hours} of videos) for 6 hours. On the other hand, PD-FGC decouples expression space on Voxceleb2 dataset (\textbf{2400 hours}) by discorelation loss for 2 weeks. The visualization in Fig.~\ref{fig:time} allows for a more intuitive comparison of the differences between the different methods concerning required training time, training data, and computational arithmetic.

\subsection{More Comparison with SOTA Face Reenactment Methods}

\begin{table}[t]
  \resizebox{\linewidth}{!}{

  \begin{tabular}{@{}l|cccccc@{}}
    \toprule
    Method/Metric & PSNR$\uparrow$ & SSIM$\uparrow$ & LPIPS$\downarrow$ & $\mathcal{L}_1 \downarrow$ & AKD$\downarrow$ & AED$\downarrow$ \\
    \midrule
    PIRenderer~\cite{ren2021pirenderer} & 22.13 & 0.72 & 0.22 & 0.053 & 2.24 & 0.032 \\
    OSFV~\cite{wang2021one} & 23.29 & 0.74 & 0.17 & 0.037 & 1.83 & 0.025 \\
    LIA~\cite{wang2021latent} & 24.75 & 0.77 & 0.16 & 0.036 & 1.88 & 0.019 \\
    DaGAN~\cite{hong2022depth} & 23.21 & 0.74 & 0.16 & 0.041 & 1.93 & 0.023 \\
    MCNET~\cite{hong2023implicit} & 21.74 & 0.69 & 0.26 & 0.057 & 2.05 & 0.037 \\
    StyleHEAT~\cite{yin2022styleheat} & 22.15 & 0.65 & 0.25 & 0.075 & 2.95 & 0.045 \\
    VPGC~\cite{wang2023efficient} & - & - & - & - & - & - \\
    \midrule
    EDTalk++ & \textbf{26.5} & \textbf{0.85} & \textbf{0.13} & \textbf{0.031} & \textbf{1.74}& \textbf{0.017}  \\
    \bottomrule
  \end{tabular}
  }
  \caption{The quantitative results compared with SOTA face reenactment methods on HDTF dataset.}
  \label{tab:face_rea}
  \vspace{-0.1in}
\end{table}

\paragraph{Qualitative results}
We perform a comparative analysis with state-of-the-art face reenactment methods, including PIRenderer~\cite{ren2021pirenderer}, OSFV~\cite{wang2021one}, LIA~\cite{wang2021latent}, DaGAN~\cite{hong2022depth}, MCNET~\cite{hong2023implicit}, StyleHEAT~\cite{yin2022styleheat}, and VPGC~\cite{wang2023efficient}, where VPGC is a person-specific model. Given that the compared methods are not specifically trained on emotional datasets, we conduct comparisons using videos with and without emotion, the results of which are presented in the Supplementary Video (5:11-5:44). Our method demonstrates superior performance in terms of face reenactment.

\paragraph{Quantitative results}
We additionally offer extensive quantitative comparisons regarding: (1) Generated video quality assessed through PSNR and SSIM. (2) Reconstruction faithfulness evaluated using LPIPS and $\mathcal{L}_1$ norms. (3) Semantic consistency measured by average keypoint distance (AKD) and average Euclidean distance (AED). The quantitative results on the HDTF dataset are outlined in Tab.~\ref{tab:face_rea}, showcasing the superior performance of our EDTalk++ method. Note that since VPGC is a person-specific model, it cannot be generalized on identities in HDTF dataset.

\subsection{More Interesting Results}
\label{subsec:robustness}

\paragraph{Robustness}
Our method demonstrates robustness across out-of-domain portraits, encompassing real human subjects, paintings, sculptures, and images generated by Stable Diffusion~\cite{rombach2021highresolution}. Moreover, our approach exhibits generalizability to various audio inputs, including songs, diverse languages (English, French, German, Italian, Japanese, Korean, Spanish, Chinese), and noisy audio. Please refer to the Supplementary Video (5:49-7:52) for the better visualization of these results.

\begin{figure}[t]
  \includegraphics[width=0.95\linewidth]{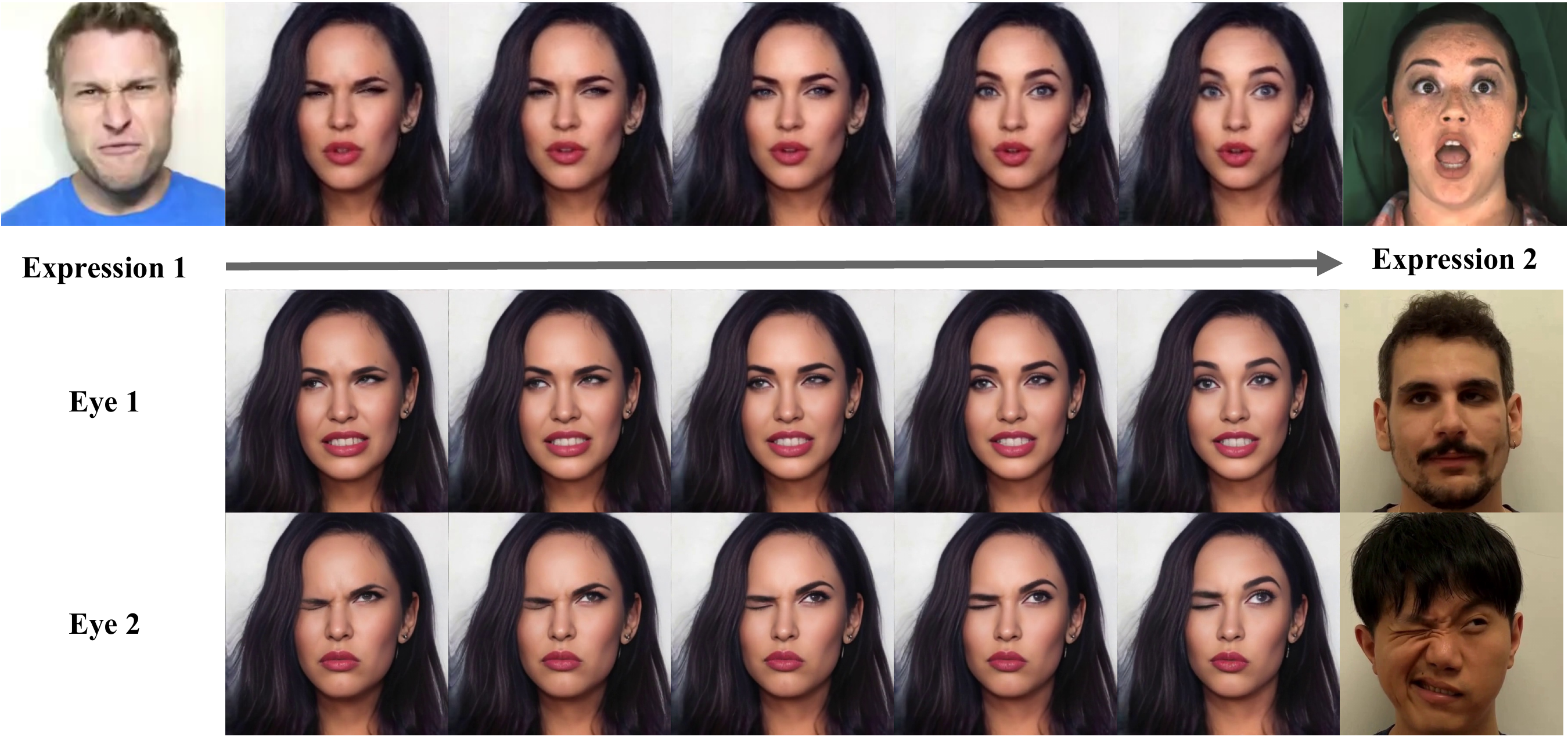}
    \caption{The results of expression manipulation.}
    \label{fig:expression_manipulation}
    \vspace{-0.2in}
\end{figure}

\paragraph{Expression Manipulation}
We accomplish expression manipulation by interpolating between expression weights $W^{ex}$ of the expression bank $B^{ex}$, which are extracted from any two distinct expression reference clips, using the following equation:
\begin{equation}
    W^{ex} = \alpha W^{ex}_1+(1-\alpha)W^{ex}_2,
\end{equation}
where $W^{ex}_1$ and $W^{ex}_2$ represent expression weights extracted from two emotional clips, while $\alpha$ denotes the interpolation weight. Fig.~\ref{fig:expression_manipulation} illustrates an example of expression manipulation generated by our EDTalk++. In this example, we successfully transition from $Expression 1$ to $Expression 2$ by varying the interpolation weight $\alpha$. This demonstrates the effectiveness of our $EXLN$ module in accurately capturing the expression of the provided clip, as discussed in the main paper.

\begin{figure}[t] 
  \includegraphics[width=\linewidth]{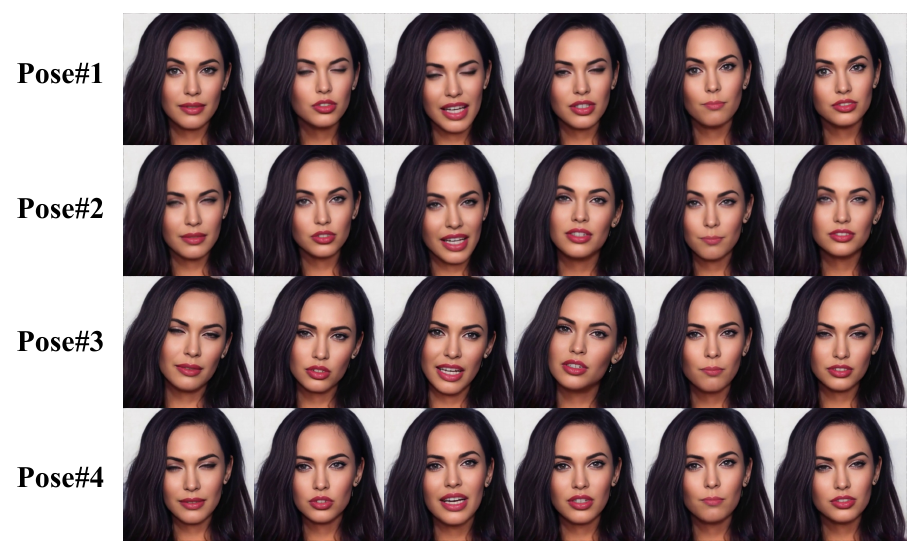} 
  \caption{The results of generated head poses.}
  \vspace{-0.2in}
  \label{fig:pose1}
\end{figure}

\paragraph{Probabilistic Pose\&Eye Generation}
Thanks to the distribution in the diffusion model of the Audio2Pose\&Eye module, we are able to sample diverse and realistic head poses and eye movements from it. As shown in Fig.~\ref{fig:pose1}, by passing the same inputs through our EDTalk++, our method synthesizes various yet natural head motions and eye movements while preserving the expression and mouth shape unchanged.  

\begin{figure*}[t]
  \centering
  \includegraphics[width=0.95\linewidth]{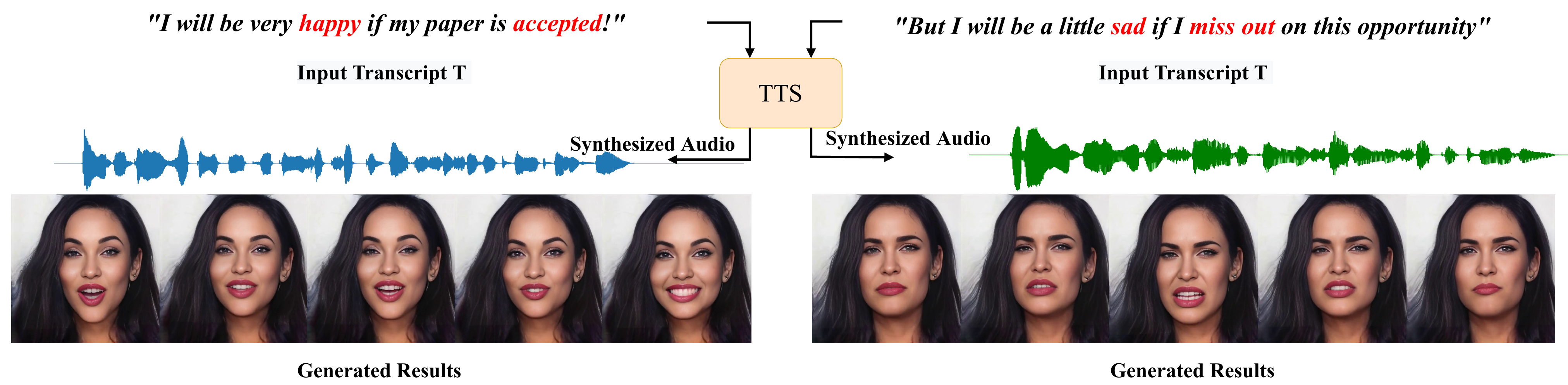}
    \caption{The results of semantically-aware expression generation.}
    \label{fig:audio2exp}
\end{figure*}

\paragraph{Semantically-Aware Expression Generation}
We input two transcripts into a Text-To-Speech (TTS) system to synthesize two audio clips. These audios, along with their respective transcripts, are then fed into our Audio-to-Motion module to generate talking face videos. The results of semantically-aware expression generation are depicted in Fig.~\ref{fig:audio2exp}, showcasing our method's ability to accurately generate expressions corresponding to the transcripts (left: happy; right: sad). Additionally, in the Supplementary Video (3:08-3:21), we provide further results where expressions are inferred directly from audio.

\begin{figure*}[t]
  \centering
  \includegraphics[width=0.95\linewidth]{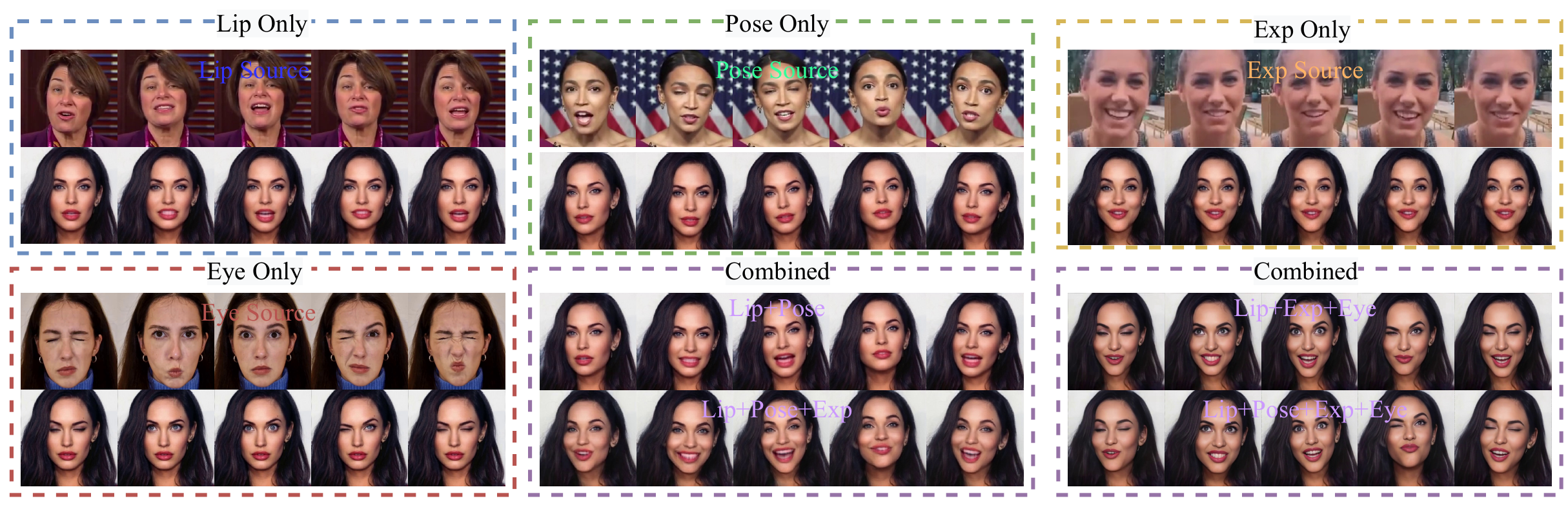}
    \caption{The results of individual control over mouth shape, head pose, emotional expression and combined facial dynamics.}
    \label{fig:base1}
    \vspace{-0.1in}
\end{figure*}

\paragraph{Motion Direction Controlled by Base}
We initially present the results showcasing individual control over mouth shape, head pose, and emotional expression in Fig.~\ref{fig:base1}. Specifically, by feeding our EDTalk++ with an identity source and various driving sources (first row of each part), our method generates corresponding disentangled outcomes in the second row. Subsequently, we integrate these individual facial motions into full emotional talking head videos with synchronized lip movements, head gestures, and emotional expressions. It's worth noting that our method facilitates the combination of any two facial parts, such as `expression+lip', `expression+pose', etc. An example of `lip+pose' and `lip+pose+eye' are shown in the second row of Fig.~\ref{fig:base1}.

\begin{figure*}[t]
  \centering
  \includegraphics[width=0.95\linewidth]{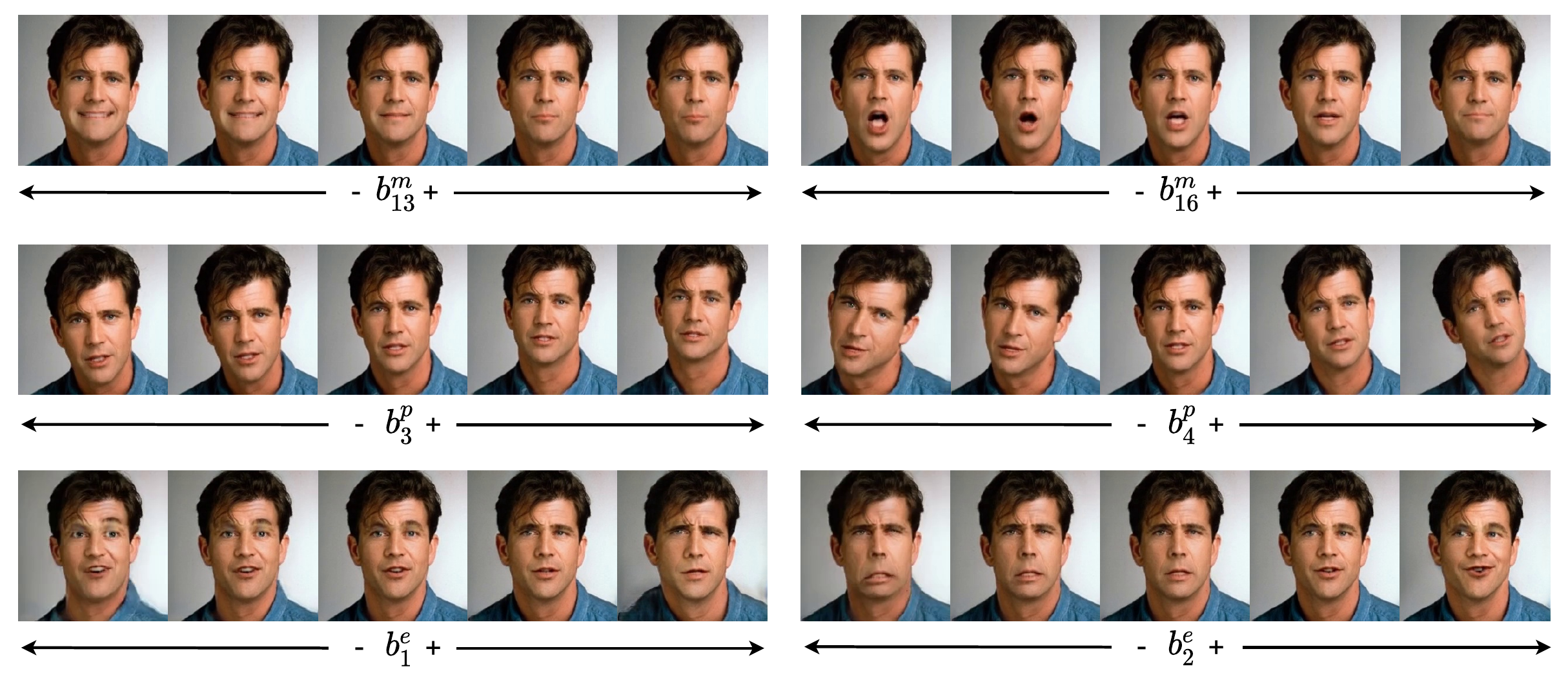}
    \caption{Motion direction controlled by each base.}
    \label{fig:base2}
\end{figure*}

We are also intrigued by understanding how each base in the banks influences motion direction. Consequently, we manipulate only a specific base $b^*_i$ and repeat the setup. The results, as depicted in Fig.~\ref{fig:base2}, indicate that the bases hold semantic significance for fundamental visual transformations such as mouth opening/closing, head rotation, and happiness/sadness/anger.

\subsection{Ablation Study}

\begin{figure}[t] 
  \includegraphics[width=\linewidth]{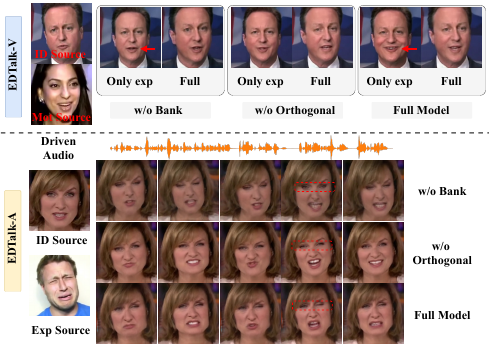} 
  \caption{Ablation results.}
  \label{fig:ablation_audio}
\end{figure}

\paragraph{Latent space} To analyze the contributions of our key designs on obtaining the disentangled latent spaces, we conduct an ablation study with two variants: (1) remove base banks (\textbf{w/o Bank}). (2) remove orthogonal constraint (\textbf{w/o Orthogonal}). Fig.~\ref{fig:ablation_audio} presents our ablation study results on video-driven and audio-driven settings, respectively. Since \textbf{w/o Bank} struggles to decouple different latent spaces, \textit{only exp} fails to extract the emotional expression. Additionally, without the visual information stored in banks, the quality of the generated full frame is poor. Although \textbf{w/o Orthogonal} improves the image quality through vision-rich banks, due to the lack of orthogonality constraints on the base, it interferes with different spaces, resulting in less obvious generated emotions. The Full Model achieves the best performance in both aspects.

\paragraph{Bank size}
In this section, we perform a series of experiments on the MEAD dataset to explore the impact of base number selection on final performance. Specifically, we vary the base number of the Mouth Bank $B^m$ and Expression Bank $B^{ex}$ across values of 5, 10, 20, and 40, respectively. The quantitative results are provided in Tab.~\ref{tab:ablation_study}, where we observe the best performance when utilizing 20 bases in $B^m$ and 10 bases in $B^{ex}$.

\begin{table*}[t]
	\centering
	\resizebox{\linewidth}{!}{
	\begin{tabular}{l|ccccc|ccccc}
		\toprule
		\multicolumn{1}{c}{\multirow{2}[4]{*}{\textbf{Method}}} & \multicolumn{5}{c}{Mouth Bank $B^m$} & \multicolumn{5}{c}{Expression Bank $B^{ex}$}\\
		\cmidrule(lr){2-6}  \cmidrule(lr){7-11}  \multicolumn{1}{c}{} & \multicolumn{1}{c}{PSNR$\uparrow$}& \multicolumn{1}{c}{SSIM$\uparrow$} & \multicolumn{1}{c}{M/F-LMD$\downarrow$}  & \multicolumn{1}{c}{$\text{Sync}_\text{conf}\uparrow$}& \multicolumn{1}{c}{$\text{Acc}_\text{emo}\uparrow$}& \multicolumn{1}{c}{PSNR$\uparrow$}& \multicolumn{1}{c}{SSIM$\uparrow$} & \multicolumn{1}{c}{M/F-LMD$\downarrow$}  & \multicolumn{1}{c}{$\text{Sync}_\text{conf}\uparrow$}& \multicolumn{1}{c}{$\text{Acc}_\text{emo}\uparrow$}
		\\

		\midrule
		5 & 20.39 & 0.69 & 2.02/1.67 & 6.35 & 63.53 & 21.54 & 0.70 & 1.60/1.35 & 8.27 & 53.26    \\
		10 & 21.45 & \textbf{0.72} & 1.65/1.33 & 7.89 & 65.74 & \textbf{22.42} & \textbf{0.75} & \textbf{1.21}/\textbf{1.19} & \textbf{8.04} & \textbf{68.21}   \\
		20 & \textbf{22.42} & \textbf{0.75} & \textbf{1.21}/\textbf{1.19} & \textbf{8.04} & 68.21 & 21.37 & \textbf{0.72} & 1.64/1.46 & 8.23 & 61.34   \\
		40 & 20.79 & 0.71 & 1.65/1.48 & 7.62 & 63.12 & 21.41 & 0.71 & 1.68/1.42 & 8.16 & 59.65  \\
		\bottomrule
	
	\end{tabular}%
	}
	\caption{Ablation study on the number of base. }
	\label{tab:ablation_study}%
\end{table*}%

\section{Discussion}

\paragraph{Novelty}
Our approach is efficient thanks to the constraints we impose on the latent spaces (\textcolor{cyan}{requirement (a), (b)}). Based on these requirements, we propose a simple and easy-to-implement framework and training strategy. This does not require large amounts of training time, training data, and computational resources. However, it does not indicate a lack of innovation in our approach. Quite the contrary, in an age where computational power reigns, our aim is to propose an efficient strategy that attains state-of-the-art performance with minimal computational resources, eschewing complex network architectures or training gimmicks. We aspire for our method to offer encouragement and insight to researchers operating within resource-constrained environments, presented in a simple and elegant manner!

\begin{figure}[t] 
  \includegraphics[width=\linewidth]{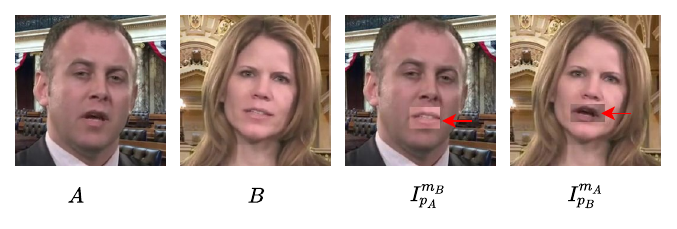} 
  \caption{Examples of synthesized images. $I^{m_B}_{p_A}$ refers to image $A$ with the mouth of $B$, and vice versa.}
  \label{fig:color}
\end{figure}

\paragraph{Potential Worries about Cross-Recontruction Disentanglement Module} We notice that there exist some color artifacts in synthesized images (pointed by red arrows in Fig.~\ref{fig:color}). However, we argue that these artifacts do not significantly impact performance and provide a detailed analysis to support this claim. \textbf{(1)} Our Encoder $E$ and Generator $G$ are pretrained in a similar setting as LIA~\cite{wang2021latent}, using a dataset collected from various sources with diverse identities, backgrounds, and motions. This diversity results in richness and colorfulness in each frame, making the Encoder $E$ robust to different input images. We have verified this robustness in our experiments (see Sec.~\ref{subsec:robustness}). Therefore, despite the presence of artifacts, the Encoder $E$ can effectively process synthetic images. \textbf{(2)} During the training process, we employ not only cross-reconstruction but also self-reconstruction loss ($\mathcal{L}_{self}$) on images without mouth replacement. This loss makes the training data contain not only synthesized images but also a large number of unmodified images, thereby preventing performance degradation. We have also confirmed the contribution of self-reconstruction through our ablation study.



\paragraph{Limitation}
While our current work has made significant strides, it also possesses certain limitations. Firstly, our method currently overlooks the influence of emotion on head pose, which represents a meaningful yet unexplored task. Unfortunately, the existing emotional MEAD dataset~\cite{wang2020mead} maintains consistent head poses across emotions, making it challenging to model the impact of emotion on pose. However, once relevant datasets become available, our approach can readily be extended to incorporate the influence of emotion on head pose by introducing emotion labels $e$ as an additional conditioning factor, as depicted in Eq. (13): $\hat{W}^p = \varphi_{p,ey}(z, f^a_t, e)$. Secondly, Second, since our method is based on a warp-based GAN model, it may produce visual artifacts under extreme poses, such as profile views, due to the limitations of 2D warping. In future work, we plan to incorporate depth or 3D structural information to alleviate such issues and improve robustness under challenging view angles.


\paragraph{Ethical considerations}
Our approach is geared towards generating talking face animations with individual facial control, which holds promise for various applications such as entertainment and filmmaking. However, there is a potential for malicious misuse of this technology on social media platforms, leading to negative societal implications. Despite significant advancements in deepfake detection research~\cite{LucyChai2020WhatMF, DavidGuera2018DeepfakeVD, arandjelovic2017look, YipinZhou2021JointAD}, there is still room for improvement in detection accuracy, particularly with the availability of more diverse and comprehensive datasets. In this regard, we are pleased to offer our talking face results, which can contribute to enhancing detection algorithms to better handle increasingly sophisticated scenarios.

\section{Conclusion}
This paper introduces EDTalk++, a novel system designed to efficiently disentangle facial components into latent spaces, enabling fine-grained control for talking head synthesis. The core insight is to represent each space with orthogonal bases stored in dedicated banks. We propose an efficient training strategy that autonomously allocates spatial information to each space, eliminating the necessity for external or prior structures. By integrating these spaces, we enable audio-driven talking head generation through a lightweight Audio-to-Motion module. Experiments showcase the superiority of our method in achieving disentangled and precise control over diverse facial motions.


\bibliographystyle{IEEEtran}
\bibliography{main}

\newpage


\vfill

\end{document}